\documentclass[final,5p,times,twocolumn]{elsarticle}
\usepackage{amsmath,amsfonts}
\usepackage{algorithmic}
\usepackage{algorithm}
\usepackage{array}
\usepackage[caption=false,font=normalsize,labelfont=sf,textfont=sf]{subfig}
\usepackage{booktabs}
\usepackage{textcomp}
\usepackage{stfloats}
\usepackage{url}
\usepackage{verbatim}
\usepackage{graphicx}
\usepackage{multirow}
\usepackage{amssymb}
\usepackage{stfloats}
\usepackage{diagbox}
\usepackage{amsmath,amsfonts}
\usepackage{amsthm}
\usepackage[normalem]{ulem}
\useunder{\uline}{\ul}{}
\newtheorem{definition}{Definition}[section]

\usepackage{xcolor}
\usepackage{graphicx}
\newcommand{\revision}{\textcolor{black}}

\usepackage{hyperref}
\hypersetup{
    colorlinks=true,
    allcolors=blue,
    pdftitle={Control-flow anomaly detection by process mining-based feature extraction and dimensionality reduction},
    pdfpagemode=FullScreen,
}

\def\nomove{\gg}

\journal{Knowledge-Based Systems}

\begin{document}
\begin{frontmatter}

\title{Control-flow anomaly detection by process mining-based\\feature extraction and dimensionality reduction}

\author[federicoii]{Francesco Vitale}
\author[rwth]{Marco Pegoraro}
\author[rwth]{Wil M. P. van der Aalst}
\author[federicoii]{Nicola Mazzocca}
\affiliation[federicoii]{organization={University of Naples Federico II},addressline={Via Claudio, 21}, city={Naples}, postcode={80125}, state={Campania}, country={Italy}}
\affiliation[rwth]{organization={RWTH Aachen University},addressline={Ahornstr. 55}, city={Aachen}, postcode={52074}, state={Nordrhein-Westfalen}, country={Germany}}

\begin{abstract}
The business processes of organizations may deviate from normal control flow due to disruptive anomalies, including unknown, skipped, and wrongly-ordered activities. To identify these control-flow anomalies, \revision{process mining} can check control-flow correctness against a reference process model through \revision{conformance checking}, an explainable set of algorithms that allows linking any deviations with model elements. However, the effectiveness of \revision{conformance checking}-based techniques is negatively affected by noisy event data and low-quality process models. To address these shortcomings and support the development of competitive and explainable \revision{conformance checking}-based techniques for control-flow anomaly detection, we propose a novel \revision{process mining}-based feature extraction approach with alignment-based \revision{conformance checking}. This variant aligns the deviating control flow with a reference process model; the resulting alignment can be inspected to extract additional statistics such as the number of times a given activity caused mismatches. We integrate this approach into a flexible and explainable framework for developing techniques for control-flow anomaly detection. The framework combines \revision{process mining}-based feature extraction and dimensionality reduction to handle high-dimensional feature sets, achieve detection effectiveness, and support explainability. The results show that the framework techniques implementing our approach outperform the baseline \revision{conformance checking}-based techniques while maintaining the explainable nature of \revision{conformance checking}. We also provide an explanation of why existing \revision{conformance checking}-based techniques may be ineffective. Finally, the results indicate that detection effectiveness is not solely dependent on the specific framework technique used, as no one-size-fits-all \revision{process mining}-based feature extraction approach is suitable for all the synthetic and real-world datasets. 
\end{abstract}


\end{frontmatter}

\section{Introduction}
\label{sec:introduction}
\begin{figure}[!t]
\centering
\includegraphics[width=0.9\columnwidth]{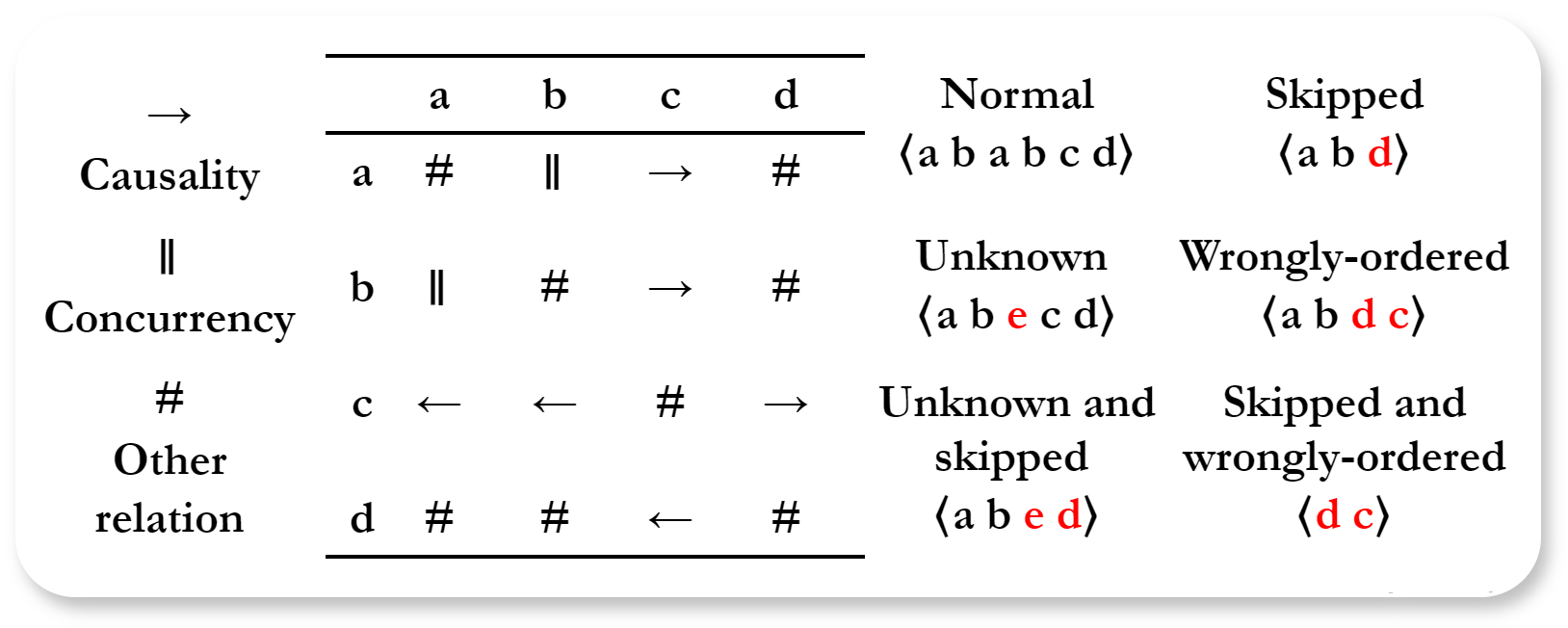}
\caption{An example event log footprint with six traces, of which five exhibit control-flow anomalies.}
\label{FP_ANOMALIES}
\end{figure}

The business processes of organizations are experiencing ever-increasing complexity due to the large amount of data, high number of users, and high-tech devices involved \cite{martin2021pmopportunitieschallenges, beerepoot2023biggestbpmproblems}. This complexity may cause business processes to deviate from normal control flow due to unforeseen and disruptive anomalies \cite{adams2023proceddsriftdetection}. These control-flow anomalies manifest as unknown, skipped, and wrongly-ordered activities in the traces of event logs monitored from the execution of business processes \cite{ko2023adsystematicreview}. For the sake of clarity, let us consider an illustrative example of such anomalies. Figure \ref{FP_ANOMALIES} shows a so-called event log footprint, which captures the control flow relations of four activities of a hypothetical event log. In particular, this footprint captures the control-flow relations between activities \texttt{a}, \texttt{b}, \texttt{c} and \texttt{d}. These are the causal ($\rightarrow$) relation, concurrent ($\parallel$) relation, and other ($\#$) relations such as exclusivity or non-local dependency \cite{aalst2022pmhandbook}. In addition, on the right are six traces, of which five exhibit skipped, wrongly-ordered and unknown control-flow anomalies. For example, $\langle$\texttt{a b d}$\rangle$ has a skipped activity, which is \texttt{c}. Because of this skipped activity, the control-flow relation \texttt{b}$\,\#\,$\texttt{d} is violated, since \texttt{d} directly follows \texttt{b} in the anomalous trace.

\subsection{Control-flow anomaly detection}
Control-flow anomaly detection techniques aim to characterize the normal control flow from event logs and verify whether these deviations occur in new event logs \cite{ko2023adsystematicreview}. To develop control-flow anomaly detection techniques, \revision{process mining} has seen widespread adoption owing to process discovery and \revision{conformance checking}. On the one hand, process discovery is a set of algorithms that encode control-flow relations as a set of model elements and constraints according to a given modeling formalism \cite{aalst2022pmhandbook}; hereafter, we refer to the Petri net, a widespread modeling formalism. On the other hand, \revision{conformance checking} is an explainable set of algorithms that allows linking any deviations with the reference Petri net and providing the fitness measure, namely a measure of how much the Petri net fits the new event log \cite{aalst2022pmhandbook}. Many control-flow anomaly detection techniques based on \revision{conformance checking} (hereafter, \revision{conformance checking}-based techniques) use the fitness measure to determine whether an event log is anomalous \cite{bezerra2009pmad, bezerra2013adlogspais, myers2018icsadpm, pecchia2020applicationfailuresanalysispm}. 

The scientific literature also includes many \revision{conformance checking}-independent techniques for control-flow anomaly detection that combine specific types of trace encodings with machine/deep learning \cite{ko2023adsystematicreview, tavares2023pmtraceencoding}. Whereas these techniques are very effective, their explainability is challenging due to both the type of trace encoding employed and the machine/deep learning model used \cite{rawal2022trustworthyaiadvances,li2023explainablead}. Hence, in the following, we focus on the shortcomings of \revision{conformance checking}-based techniques to investigate whether it is possible to support the development of competitive control-flow anomaly detection techniques while maintaining the explainable nature of \revision{conformance checking}.
\begin{figure}[!t]
\centering
\includegraphics[width=\columnwidth]{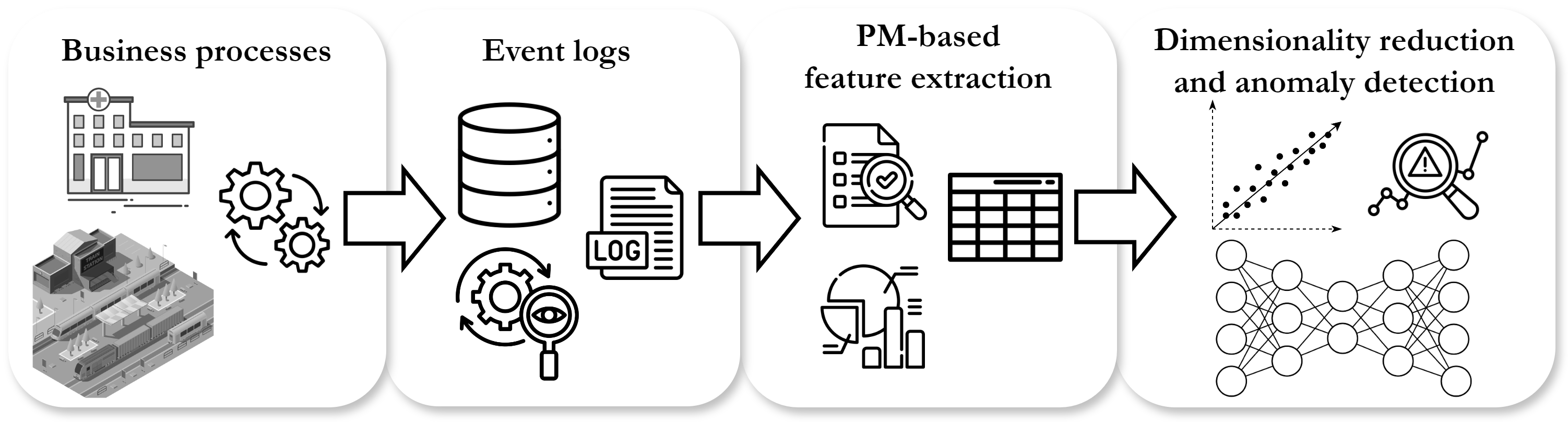}
\caption{A high-level view of the proposed framework for combining \revision{process mining}-based feature extraction with dimensionality reduction for control-flow anomaly detection.}
\label{HIGH_LEVEL_VIEW}
\end{figure}

\subsection{Shortcomings of \revision{conformance checking}-based techniques}
Unfortunately, the detection effectiveness of \revision{conformance checking}-based techniques is affected by noisy data and low-quality Petri nets, which may be due to human errors in the modeling process or representational bias of process discovery algorithms \cite{bezerra2013adlogspais, pecchia2020applicationfailuresanalysispm, aalst2016pm}. Specifically, on the one hand, noisy data may introduce infrequent and deceptive control-flow relations that may result in inconsistent fitness measures, whereas, on the other hand, checking event logs against a low-quality Petri net could lead to an unreliable distribution of fitness measures. Nonetheless, such Petri nets can still be used as references to obtain insightful information for \revision{process mining}-based feature extraction, supporting the development of competitive and explainable \revision{conformance checking}-based techniques for control-flow anomaly detection despite the problems above. For example, a few works outline that token-based \revision{conformance checking} can be used for \revision{process mining}-based feature extraction to build tabular data and develop effective \revision{conformance checking}-based techniques for control-flow anomaly detection \cite{singh2022lapmsh, debenedictis2023dtadiiot}. However, to the best of our knowledge, the scientific literature lacks a structured proposal for \revision{process mining}-based feature extraction using the state-of-the-art \revision{conformance checking} variant, namely alignment-based \revision{conformance checking}.

\subsection{Contributions}
We propose a novel \revision{process mining}-based feature extraction approach with alignment-based \revision{conformance checking}. This variant aligns the deviating control flow with a reference Petri net; the resulting alignment can be inspected to extract additional statistics such as the number of times a given activity caused mismatches \cite{aalst2022pmhandbook}. We integrate this approach into a flexible and explainable framework for developing techniques for control-flow anomaly detection. The framework combines \revision{process mining}-based feature extraction and dimensionality reduction to handle high-dimensional feature sets, achieve detection effectiveness, and support explainability. Notably, in addition to our proposed \revision{process mining}-based feature extraction approach, the framework allows employing other approaches, enabling a fair comparison of multiple \revision{conformance checking}-based and \revision{conformance checking}-independent techniques for control-flow anomaly detection. Figure \ref{HIGH_LEVEL_VIEW} shows a high-level view of the framework. Business processes are monitored, and event logs obtained from the database of information systems. Subsequently, \revision{process mining}-based feature extraction is applied to these event logs and tabular data input to dimensionality reduction to identify control-flow anomalies. We apply several \revision{conformance checking}-based and \revision{conformance checking}-independent framework techniques to publicly available datasets, simulated data of a case study from railways, and real-world data of a case study from healthcare. We show that the framework techniques implementing our approach outperform the baseline \revision{conformance checking}-based techniques while maintaining the explainable nature of \revision{conformance checking}.

In summary, the contributions of this paper are as follows.
\begin{itemize}
    \item{
        A novel \revision{process mining}-based feature extraction approach to support the development of competitive and explainable \revision{conformance checking}-based techniques for control-flow anomaly detection.
    }
    \item{
        A flexible and explainable framework for developing techniques for control-flow anomaly detection using \revision{process mining}-based feature extraction and dimensionality reduction.
    }
    \item{
        Application to synthetic and real-world datasets of several \revision{conformance checking}-based and \revision{conformance checking}-independent framework techniques, evaluating their detection effectiveness and explainability.
    }
\end{itemize}

The rest of the paper is organized as follows.
\begin{itemize}
    \item Section \ref{sec:related_work} reviews the existing techniques for control-flow anomaly detection, categorizing them into \revision{conformance checking}-based and \revision{conformance checking}-independent techniques.
    \item Section \ref{sec:abccfe} provides the preliminaries of \revision{process mining} to establish the notation used throughout the paper, and delves into the details of the proposed \revision{process mining}-based feature extraction approach with alignment-based \revision{conformance checking}.
    \item Section \ref{sec:framework} describes the framework for developing \revision{conformance checking}-based and \revision{conformance checking}-independent techniques for control-flow anomaly detection that combine \revision{process mining}-based feature extraction and dimensionality reduction.
    \item Section \ref{sec:evaluation} presents the experiments conducted with multiple framework and baseline techniques using data from publicly available datasets and case studies.
    \item Section \ref{sec:conclusions} draws the conclusions and presents future work.
\end{itemize}
\section{Related work}
\label{sec:related_work}
The large body of research on control-flow anomaly detection in event logs sees a wide range of data-driven approaches \cite{ko2023adsystematicreview}. In particular, the scientific literature developed techniques for control-flow anomaly detection that may or may not use \revision{conformance checking}. To distinguish between those that use \revision{conformance checking} and those that do not, we refer to them as \revision{conformance checking}-based techniques and \revision{conformance checking}-independent techniques.

\subsection{\revision{Conformance checking}-independent techniques}
\revision{Conformance checking}-independent techniques use specific types of trace encodings to extract features from event logs, such as one-hot encoding and word2vec \cite{ko2023adsystematicreview}. In particular, trace encodings may extract process-based statistics, such as N-grams or directly-follows relationships \cite{tavares2023pmtraceencoding}. We refer to these types of trace encodings as \textit{statistical} \revision{process mining}-based feature extraction.

\paragraph{Deep learning} A large class of \revision{conformance checking}-independent techniques rely on deep learning by feed-forward and/or recurrent neural networks, such as the autoencoder \cite{nolle2018processadautoencoders, vijayakamal2020bpaead, elaziz2023drlbpad, chinnaiah2024deepaead, kan2024aebasedelad}, long-short term memory networks \cite{du2017deeplog, lahann2022lstmadpi} and gated recurrent unit networks \cite{nolle2022binet, guan2023gruaebpad}. Some works combined these approaches, employing both the autoencoder and recurrent neural networks to exploit the ability of reconstruction-based and prediction-based anomaly detection \cite{yuan2021deeplstmaead, wang2022lstmaeaddiagnosis, guan2024wake}. Although \revision{conformance checking}-independent techniques using deep learning are very effective, their performance could differ based on the type of trace encoding \cite{tavares2021pmencodingad}. Moreover, deep learning is notoriously affected by explainability issues \cite{rawal2022trustworthyaiadvances}. To address these issues, the scientific literature aimed to combine deep learning with \revision{process mining}. Firstly, recurrent neural networks have been used to approximate the diagnostics that alignment-based \revision{conformance checking} provides; arguably, these workarounds reduce the computational cost of alignment-based \revision{conformance checking} while reaching comparable results \cite{nolle2020deepalign, boltenhagen2020costbasedlogtracesclassification}. Additionally, other approaches used deep learning to pre-process data so that \revision{process mining} could provide better performance. For example, Krajsic and Franczyk \cite{krajsic2021vaeadonlinepm} used the autoencoder to perform anomaly detection before \revision{process mining} to obtain higher-quality process models through process discovery. Similarly, Wang et al. \cite{wang2022lstmaeaddiagnosis} applied \revision{process mining} after \revision{conformance checking}-independent control-flow anomaly detection to provide diagnostics and inspect the causes of deviating control flow. These works suggest that the literature recognizes the utility of \revision{process mining} in improving the explainability of control-flow anomaly detection based on deep learning.

\paragraph{Statistics}Whereas many \revision{conformance checking}-independent techniques rely on deep learning due to its detection effectiveness, several proposals set forth statistical approaches. Mostly, these approaches rely on evaluating the similarity of new event data to a normal and interpretable statistic. For example, Li and van der Aalst \cite{li2017frameworkdeviations} encoded information from the directly follows and dependency relations, building ``profiles" by which new event data are compared to evaluate whether they are anomalous. Ko and Comuzzi \cite{ko2021statisticaladbp} extracted a statistical index termed leverage measure to use as a metric to calculate the ``anomaly score" of traces in event logs; this measure can be used to either fit a statistical distribution or build a threshold to which new traces are compared. Similarly, Luftensteiner and Praher \cite{luftensteiner2022adpmgraphs} presented an approach based on spectral analysis of the adjacency matrix built from a graph that encodes a given relation among events. The approach evaluates the difference between the second-highest eigenvalue of adjacency matrices --- i.e., the spectral gap --- of a normal event log and a new event log; if the spectral gap is too large, the new event log is anomalous. Finally, Mavroudopoulos and Gounaris \cite{manvroudopoulos2022proximitybasedtemporalad} proposed a nearest-neighbor-based approach that evaluates the distance between reference ``vector traces", i.e., trace encodings, and new vector traces using different metrics; if the distance exceeds a given threshold based on the set of normal trace encodings, the new vector trace is classified as anomalous.

The high heterogeneity of \revision{conformance checking}-independent techniques for control-flow anomaly detection makes a comprehensive and fair comparison between them a challenging task. The framework that we propose in Section \ref{sec:framework} aims to implement reconstruction-based anomaly detection, as done in many of the reviewed references \cite{nolle2018processadautoencoders, vijayakamal2020bpaead, elaziz2023drlbpad, chinnaiah2024deepaead, kan2024aebasedelad}. However, an overarching comparison of all \revision{conformance checking}-independent techniques is out of the scope of this work. Instead, as discussed below, we focus on enhancing \revision{conformance checking}-based techniques to both improve the detection performance and maintain the explainable nature of \revision{conformance checking}.

\subsection{\revision{Conformance checking}-based techniques}
\begin{figure}[!t]
\centering
\includegraphics[width=0.9\columnwidth]{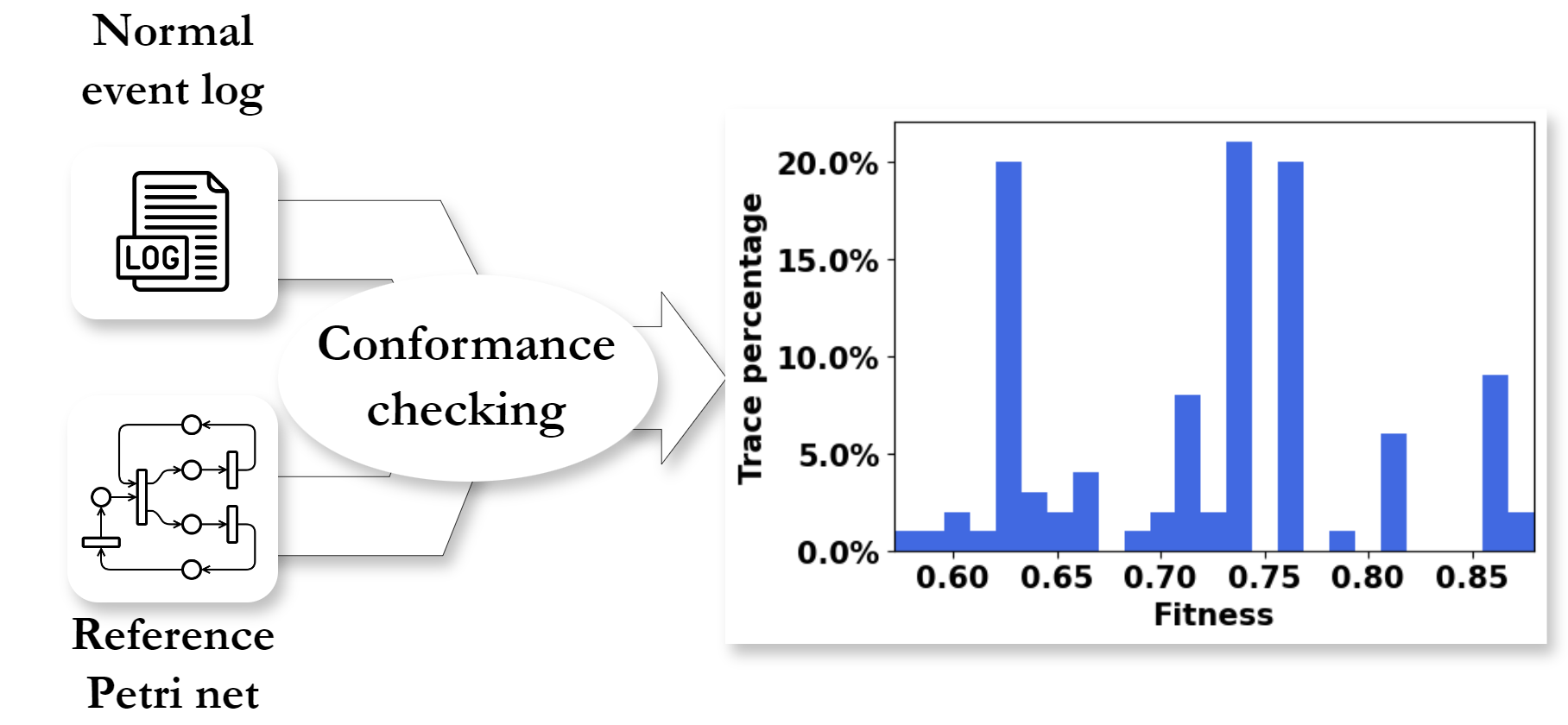}
\caption{The distribution of fitness values of case-study traces of a normal event log replayed on a reference Petri net. The inconsistent distribution of fitness values of the case-study traces makes fitness thresholding ineffective for control-flow anomaly detection.}
\label{CC_UNBALANCING}
\end{figure}

Contrary to \revision{conformance checking}-independent techniques, \revision{conformance checking}-based techniques do not employ trace encodings such as one-hot encoding and word2vec. Instead, \revision{conformance checking}-based techniques aim to seamlessly check new event logs with a reference Petri net, obtaining the fitness measure to indicate whether the event data is anomalous \cite{bezerra2009pmad, accorsi2012pmsecurityaudits, bezerra2013adlogspais, myers2018icsadpm, pecchia2020applicationfailuresanalysispm}. In addition, the fitness measure is tightly connected to the Petri net semantics, thus providing explainable ways to analyze the root cause of unfitting event logs \cite{aalst2016pm}. However, relying on fitness thresholding has shown poor detection effectiveness mostly due to noisy data and/or low-quality process models \cite{bezerra2013adlogspais, pecchia2020applicationfailuresanalysispm}. In this regard, let us consider Figure \ref{CC_UNBALANCING}, which depicts a normal event log, namely an event log containing normal traces from one of the datasets we use during the evaluation, and a reference Petri net from the same dataset. The histogram shows that a high percentage of traces have a fitness measure below 0.75. It is therefore not trivial to set a threshold based on these results. 

\paragraph{Solutions to the shortcomings}
Whereas some works disregarded the issue of noisy event logs and/or low-quality Petri nets by deeming anomalous all traces that the reference Petri net does not perfectly fit \cite{accorsi2012pmsecurityaudits, myers2018icsadpm}, other works attempted to account for these shortcomings by exploiting other information related to either the reference Petri net or control-flow diagnostics obtained by \revision{conformance checking}. For example, Bezerra et al. \cite{bezerra2009pmad} combined process discovery with \revision{conformance checking} to extract the ``most appropriate" Petri net according to both a manually tuned fitness threshold and the simplicity of the Petri net obtained by process discovery. Similarly, Bezerra et al. \cite{bezerra2013adlogspais} and Pecchia et al. \cite{pecchia2020applicationfailuresanalysispm} developed algorithms and methods to discover high-quality Petri nets with process discovery and automatically tune the fitness threshold for anomaly detection. In addition to these solutions, which somehow attempt to optimize the fitness threshold, other approaches combine \revision{conformance checking} with machine learning to supply additional information about faulty control flow, such as the number of times an activity was not allowed to execute according to the model semantics. This process leads to \revision{conformance checking} \revision{process mining}-based feature extraction, which results in tabular data that can be processed by machine/deep learning algorithms for anomaly detection \cite{sarno2020pmadbp, singh2022lapmsh, debenedictis2023dtadiiot}.

\paragraph{Our solution} To address the problem shown in Figure \ref{CC_UNBALANCING} regarding the choice of a fitness threshold, the scientific literature proposes to either a) tune an optimal fitness threshold by discovering the highest-quality process model through process discovery, or b) provide additional control-flow information obtained by \revision{conformance checking} to feed to machine learning. In this work, we follow approach b) and develop a novel \revision{process mining}-based feature extraction approach with alignment-based \revision{conformance checking}, which aligns the deviating control flow with a reference Petri net; the resulting alignment can be inspected to extract additional statistics such as the number of times a given activity caused mismatches. This allows the connection of these statistics to the reference Petri net for additional insights on the deviating control flow. We present this solution in the next section, followed by its integration into a flexible and explainable framework for developing \revision{conformance checking}-based and \revision{conformance checking}-independent techniques for control-flow anomaly detection.

\section{The proposed process mining-based feature extraction}
\label{sec:abccfe}
This section first presents the preliminaries of \revision{process mining} to establish the notation used throughout the paper. In particular, we report the standard definitions from the \revision{process mining} literature of the event log, sublog, trace, Petri net, fitness, and alignment and moves. After the necessary notation is established, the proposed \revision{process mining}-based feature extraction with alignment-based \revision{conformance checking} is developed, including the definitions of alignment-based fitness and per-activity cost. These two are used to extract alignment-based \revision{conformance checking} diagnoses, i.e., the novel \revision{process mining}-based feature extraction approach.
\subsection{Preliminaries}
Firstly, we report the definition of the event log and sublog, since they are the main inputs to \revision{process mining} algorithms.
\begin{definition}[Event log, sublog and trace]
\label{def:el}
Let us denote $\mathcal{A}$ as the universe of activities and $\mathcal{A}^*$ the universe of possible sequences of activities in $\mathcal{A}$. Hereafter, we refer to a sequence of activities as a \emph{trace} $\sigma\in\mathcal{A}^*$. An \emph{event log} $L$ is a multi-set of $k$ traces, where a multi-set is a set such that there may be two equal yet distinct elements. Given $\mathcal{B}(\mathcal{A}^*)$ the set of multi-sets over $\mathcal{A}^*$, $L\in\mathcal{B}(\mathcal{A}^*)$. A \emph{sublog} is a subset of $L$. Furthermore, we denote an \emph{$n$-tuple of event logs} $\mathcal{L}=(\mathcal{L}(1),\mathcal{L}(2),\dots, \mathcal{L}(n))$ as $\mathcal{L}\in(\mathcal{B}(\mathcal{A^*}))^n$. 
\end{definition}
An event log can be processed by process discovery algorithms to build a process model. This process model captures a set of possible control flows based on the relationships among activities within the traces of the event log. There is a plethora of formalisms to capture these relationships. As mentioned in Section \ref{sec:introduction}, in this work, we focus on the labeled Petri net (hereafter, Petri net).
\begin{definition}[Petri net]
Let $P$ and $Tr$ be two sets of nodes of a bipartite graph such that $P\,\cap\,Tr=\emptyset$, and $F\subseteq (P\times Tr)\,\cup\,(Tr\times P)$ a set of directed arcs. Furthermore, let $A\subseteq\mathcal{A}$ be a set of activities and $l:Tr\rightarrow A\cup\{\tau\}$ a function that associates to elements of $Tr$ either an activity of $A$ or the ``silent" label $\tau$. A Petri net $N$ is the tuple $(P,Tr,F,A,l)$, where $P$ are the places, $Tr$ the transitions, $F$ the arcs, $A$ the activities, and $l$ the labeling function. Finally, $M\in \mathcal{B}(P)$ is the (current) \emph{marking} of the Petri net, i.e., a multi-set of tokens that determine which transition is allowed to \emph{fire}. In the following, we denote $\mathcal{N}$ the universe of Petri nets. 
\end{definition}
Similar to process discovery, \revision{conformance checking} algorithms also process event logs, but they use a reference Petri net and check the traces of event logs against it to verify if the execution fits the model \cite{aalst2022pmhandbook}. The main output of \revision{conformance checking} is the fitness, which quantifies how much the process model fits the event log.

\begin{definition}[Fitness]
Let $L\in\mathcal{B}(\mathcal{A^*})$ be an event log, $\sigma\in L$ a trace, and $N\in\mathcal{N}$ a Petri net. We denote the \emph{fitness} computed by \revision{conformance checking} when replaying $\sigma$ on $N$ as $F_{\sigma, N}\in[0,1]$, whereas when replaying all traces of $L$ we denote the \emph{fitness} as $F_{L,N}\in[0,1]$.
\end{definition}

Although there are many \revision{conformance checking} variants in the literature, the state-of-the-art is alignment-based \revision{conformance checking}. This variant attempts to find the best path across the reference Petri net that most accurately approximates the traces in the event log that deviate from normal control flow. This is done by taking into account the \textit{log moves} in an event log and their \textit{alignment} to the reference Petri net by evaluating the corresponding \textit{model moves}.
\begin{definition}[Alignment and moves]
Let us denote $\nomove$ as the ``skipped" activity. Let $N\in\mathcal{N}$ be a Petri net and $\sigma\in \mathcal{A}^*$ a trace. $\sigma_L=\langle a_1,\dots,a_o\rangle\in (\mathcal{A}\cup\{\nomove\})^*$ is a (finite) sequence of log moves related to $\sigma$ if and only if $\sigma_L\setminus \{\nomove\}=\sigma$. Then, $\sigma_{N}=\langle b_1,\dots, b_o\rangle\in (\mathcal{A} \cup \{\tau\} \cup \{\nomove\})^*$ is a (finite) sequence of model moves if and only if $\sigma_{N}\setminus\{\nomove\}$ is a firing sequence of $N$, i.e. an allowed control flow. An \emph{alignment} $\gamma_{\sigma, N}$ is the two-row matrix
\begin{equation}
\gamma_{\sigma, N}=
\begin{array}{c|c|c|c}
a_1 & a_2 & \cdots & a_o \\
\hline
b_1 & b_2 & \cdots & b_o
\end{array},
\end{equation}
if for all $1 \leq i \leq o$, $(a_i, b_i) \neq (\nomove, \nomove)$, where $(a_i,b_i)$ is a \emph{move}. 
\end{definition}
\subsection{Alignment-based \revision{conformance checking} diagnoses}
\begin{table}[!t]
\centering
\caption{Alignment-based \revision{conformance checking} diagnoses obtained from replaying the $n$-tuple of event logs $\mathcal{L}$ against a Petri net $N$.}
\label{tab:CC_DIAGNOSES}
\begin{tabular}{llllll}
\hline
\textbf{Event log}            & $\boldsymbol{c_1}$       & $\boldsymbol{c_2}$       & $\cdots$ & $\boldsymbol{c_o}$       & $\boldsymbol{F_{\mathcal{L}(i),N}}$ \\ \hline
$\boldsymbol{\mathcal{L}(1)}$ & $u_{c_1,\mathcal{L}(1)}$ & $u_{c_2,\mathcal{L}(1)}$ & $\cdots$ & $u_{c_o,\mathcal{L}(1)}$ & $F_{\mathcal{L}(1),N}$              \\
$\boldsymbol{\mathcal{L}(2)}$ & $u_{c_1,\mathcal{L}(2)}$ & $u_{c_2,\mathcal{L}(2)}$ & $\cdots$ & $u_{c_o,\mathcal{L}(2)}$ & $F_{\mathcal{L}(2),N}$              \\
$\cdots$                      & $\cdots$                 & $\cdots$                 & $\cdots$ & $\cdots$                 & $\cdots$                            \\
$\boldsymbol{\mathcal{L}(n)}$ & $u_{c_1,\mathcal{L}(n)}$ & $u_{c_2,\mathcal{L}(n)}$ & $\cdots$ & $u_{c_o,\mathcal{L}(n)}$ & $F_{\mathcal{L}(n),N}$              \\ \hline
\end{tabular}
\end{table}

Our proposal aims to collect information recorded throughout alignment-based \revision{conformance checking} when replaying event logs on a reference Petri net; we term such information as \emph{alignment-based \revision{conformance checking} diagnoses}. These diagnoses involve computing the alignment-based fitness and the per-activity cost.
\begin{definition}[Alignment-based fitness]
\label{def:ab_fitness}
Let us denote $\Gamma_{\sigma, N}$ as the set of all alignments between a trace $\sigma\in\mathcal{A}^*$ and a Petri net $N\in\mathcal{N}$. Let $\delta$ be the unitary cost function for a pair of moves or an alignment. Finally, let $\gamma^w_{\sigma,N}\in\Gamma_{\sigma, N}$ be the worst-case alignment and $\gamma^*_{\sigma,N}\in\Gamma_{\sigma, N}$ the best-case alignment, i.e., the alignments with the least and most costs according to $\delta$, respectively.  The \emph{alignment-based fitness} for $\sigma$ is defined as
\begin{equation}
F_{\sigma,N}=1-\frac{\delta(\gamma^*_{\sigma,N})}{\delta(\gamma^w_{\sigma,N})}.
\end{equation}
Let $L\in\mathcal{B}(\mathcal{A^*})$ be an event log. The \emph{alignment-based fitness} for $L$ is defined as
\begin{equation}
F_{L,N}=\frac{\sum_{\sigma\in L}F_{\sigma,N}}{|L|}.
\end{equation}
\end{definition}
\begin{definition}[Per-activity cost]
\label{def:act_cost}
Let $N\in\mathcal{N}$ be a Petri net, $\sigma\in\mathcal{A}^*$ a trace, $\mathcal{C}_{\sigma,N}\subseteq\mathcal{A}$ the set of activities found either in $\sigma$ or in labeled transitions of $N$, $\delta$ the cost function for a pair of moves or an alignment, $\gamma^*_{\sigma,N}\in\Gamma_{\sigma, N}$ the best-case alignment, and $\gamma^*_{c,\sigma,N}$ the set of moves of $\gamma^*_{\sigma,N}$ such that $\forall(a,b)\in\gamma^*_{c,\sigma,N}, c=a\lor c=b$. The \emph{per-activity cost} for $\sigma$ is defined as
\begin{equation}
    u_{c,\sigma}=\sum_{(a,b)\in \gamma^*_{c,\sigma,N}}\delta((a,b)),\quad c\in \mathcal{C}_{\sigma,N}.
\end{equation}
Let $L\in\mathcal{B}(\mathcal{A}^*)$ be an event log and $\mathcal{C}_{L,N}\subseteq\mathcal{A}$ the set of activities found either in traces of $L$ or in labeled transitions of $N$. The per-activity cost for $L$ is:
\begin{equation}
    u_{c,L}=\sum_{\sigma\in L}u_{c,\sigma},\quad c\in \mathcal{C}_{L,N}.
\end{equation}
\end{definition}
\begin{definition}[Alignment-based \revision{conformance checking} diagnoses]
\label{def:ab_cc}
Let $\mathcal{L}\in(\mathcal{B}(\mathcal{A}^*))^n$ be an $n$-tuple of event logs, $N\in\mathcal{N}$ a Petri net, $\mathcal{C}_{\mathcal{L},N}=\{c_1,\dots,c_o\}\subseteq\mathcal{A}$ the set of activities found either in traces of event logs of $\mathcal{L}$ or in labeled transitions of $N$. \emph{Alignment-based \revision{conformance checking} diagnoses} for $\mathcal{L}$ is tabular data arranged as in Table \ref{tab:CC_DIAGNOSES}.
\end{definition}
\begin{figure}[!t]
\centering
\includegraphics[width=0.95\columnwidth]{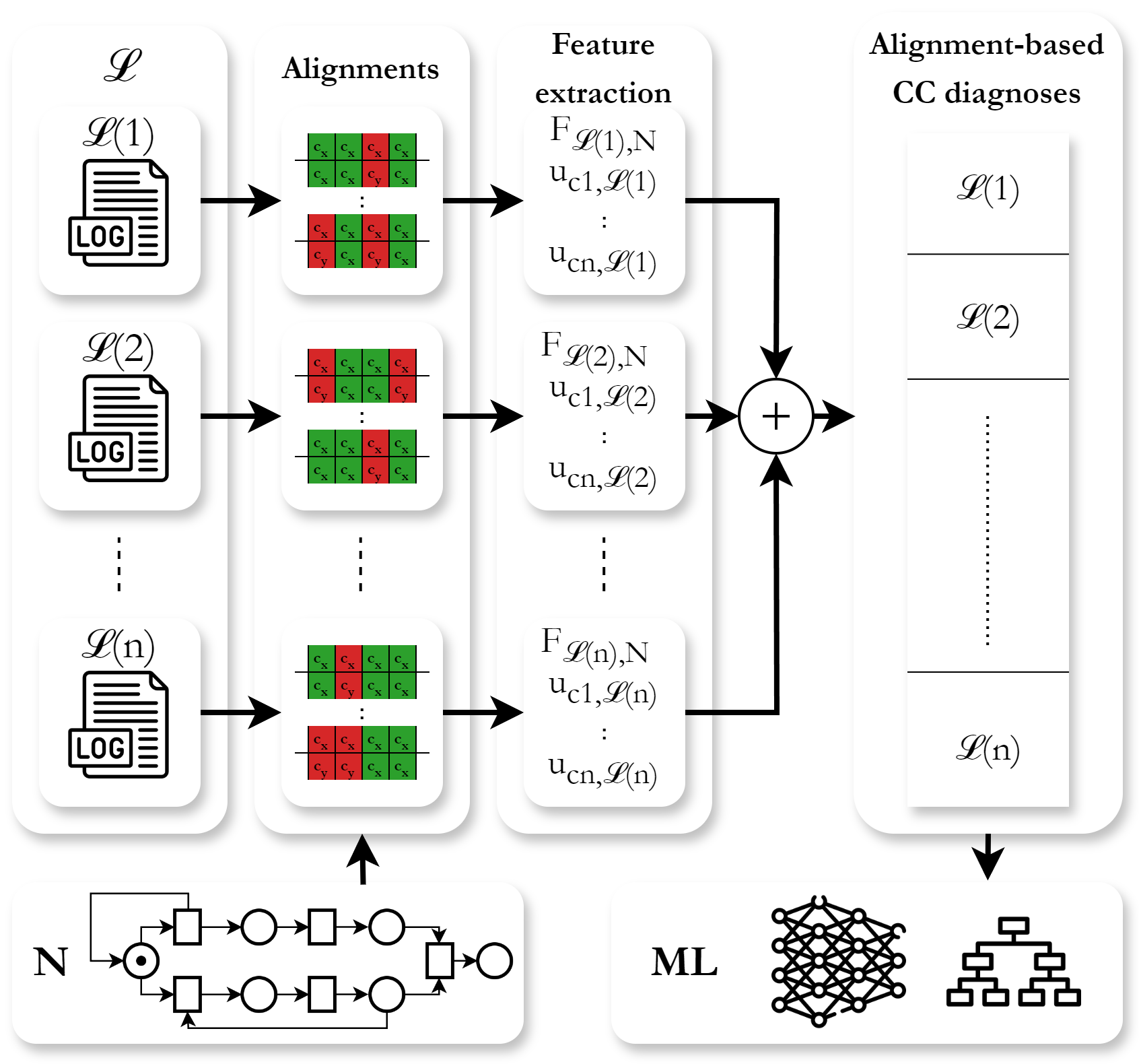}
\caption{Generation of alignment-based \revision{conformance checking} diagnoses by replaying the $n$-tuple of event logs $\mathcal{L}$ against a Petri net $N$.}
\label{AB_CC_VISUALIZATION}
\end{figure}

Figure \ref{AB_CC_VISUALIZATION} visualizes the generation of alignment-based \revision{conformance checking} diagnoses. A starting set of event logs $\mathcal{L}$ is aligned to a Petri net $N$, generating a set of alignments per event log $\mathcal{L}(1)\dots\mathcal{L}(n)$. Subsequently, alignment-based fitness $F_{\mathcal{L}(x),N}$ and per-activity costs $u_{c_1,\mathcal{L}(1)}\dots u_{c_n,\mathcal{L}(1)}$ per event log $\mathcal{L}(1)\dots\mathcal{L}(n)$ are calculated. The results are joined together to form alignment-based \revision{conformance checking} diagnoses. Finally, machine learning can handle the resulting tabular data.

\begin{figure*}[!t]
\centering
\includegraphics[width=\textwidth]{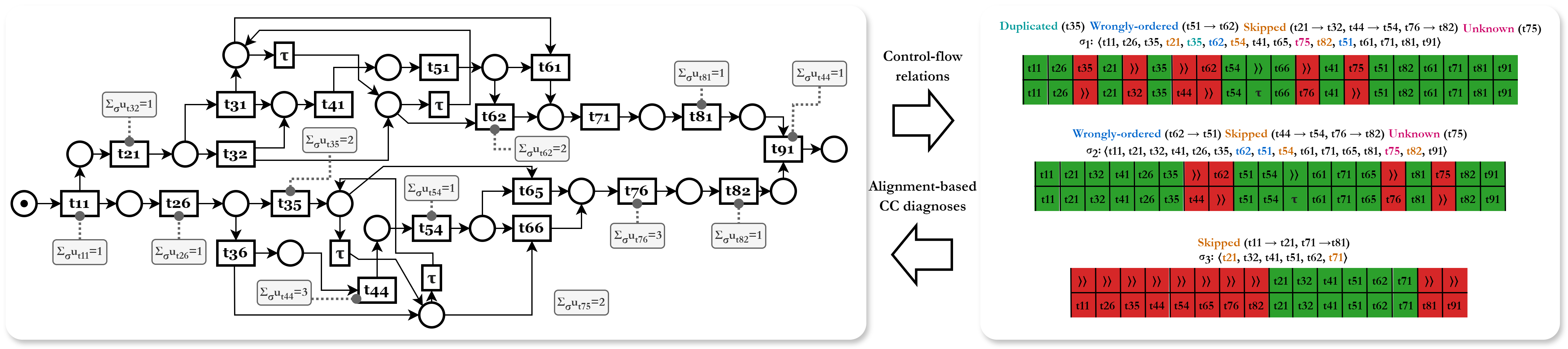}
\caption{The connection between alignment-based \revision{conformance checking} diagnoses of three traces and the reference Petri net of one of the datasets we use in Section \ref{sec:evaluation}.}
\label{PETRI_NET_TRACES}
\end{figure*}

\paragraph{Illustrative example}Figure \ref{PETRI_NET_TRACES} shows the connection between alignment-based \revision{conformance checking} diagnoses of three traces and the reference Petri net of one of the datasets we use in Section \ref{sec:evaluation}. Specifically, the reference Petri net provides control-flow relations that alignment-based \revision{conformance checking} uses to check whether traces $\sigma_1$, $\sigma_2$ and $\sigma_3$ deviate from such relations. For example, $\sigma_1$ exhibits duplicated (\texttt{t35}), wrongly-ordered (\texttt{t51}$\rightarrow$\texttt{t62}), skipped (\texttt{t21}$\rightarrow$\texttt{t32}, \texttt{t44}$\rightarrow$\texttt{t54}, \texttt{t76}$\rightarrow$\texttt{t82}), and unknown (\texttt{t75}) control-flow anomalies. As a result, this reflects in some mismatches in the corresponding alignment, highlighted by the moves (\texttt{t35},$\nomove$), ($\nomove$,\texttt{t32}), (\texttt{t62},$\nomove$), ($\nomove$,\texttt{t76}) and (\texttt{t75},$\nomove$). These mismatches can be projected onto the process model. For example, by summing all per-activity costs related to \texttt{t44}, this results in $\sum_{\sigma}u_{t44,\sigma}=3$.

\section{Framework}
\label{sec:framework}
\subsection{Problem statement}
The framework combines \revision{process mining}-based feature extraction and dimensionality reduction to allow the flexible development of several \revision{conformance checking}-based and \revision{conformance checking}-independent techniques for control-flow anomaly detection. This enables a fair comparison of multiple techniques, which may include our proposed \revision{process mining}-based feature extraction approach using alignment-based \revision{conformance checking}. 

\paragraph{\revision{Framework guidelines}}
Following Chandola et al.'s guidelines \cite{chandola2009ad} for developing anomaly detection techniques, which include defining the nature of data, type of anomalies, and the research areas employed, the framework:
\begin{itemize}
    \item Handles the event logs of business processes (see Definition \ref{def:el}) and address control-flow anomalies such as unknown, wrongly-ordered, and skipped activities;
    \item Uses \revision{conformance checking} and other types of trace encodings for \revision{process mining}-based feature extraction and generate tabular data from event logs;
    \item Employs dimensionality reduction with the tabular data resulting from training and validation event logs to build a classifier, and subsequently use the classifier to discriminate normal and anomalous event logs.
\end{itemize}

\paragraph{\revision{Scenario type}}
\revision{The type of data used during the training phase of anomaly detection techniques is highly dependent on the scenario type. For example, whereas supervised scenarios have plenty of data with accurate labels for normal and anomalous data, unsupervised scenarios involve unlabeled data only and are effective when anomalous data are infrequent \cite{chandola2009ad}. There is an intermediate scenario, namely the weakly-supervised scenario. This scenario brings the benefit of having loads of labeled normal data and a limited amount of anomalous data, and is common in business process monitoring \cite{zhang2021noisyfd, zhao2023admoe, guan2024wake}. By using only normal data during the training phase, the technique is tuned to accurately discriminate those patterns that are consistent with normal behavior, and to detect \emph{any} deviations from it. Due to weak supervision being a common scenario in business process monitoring, and the potential benefits of tailoring the anomaly detection technique to normal data only, the proposed framework deals with the weakly-supervised scenario. Hence, the framework includes only normal event logs during training and retains anomalous event logs in the test set. However, it is worth noting that weak supervision means the data may contain inaccuracies and incomplete information \cite{zhou2022weaklysupervisedaes, jiang2023weaklysupervisedad}. Consequently, noisy event logs and low-quality Petri nets can lead to issues such as those shown in Figure \ref{CC_UNBALANCING}.}

\begin{figure*}[t]
\centering
\includegraphics[width=\textwidth]{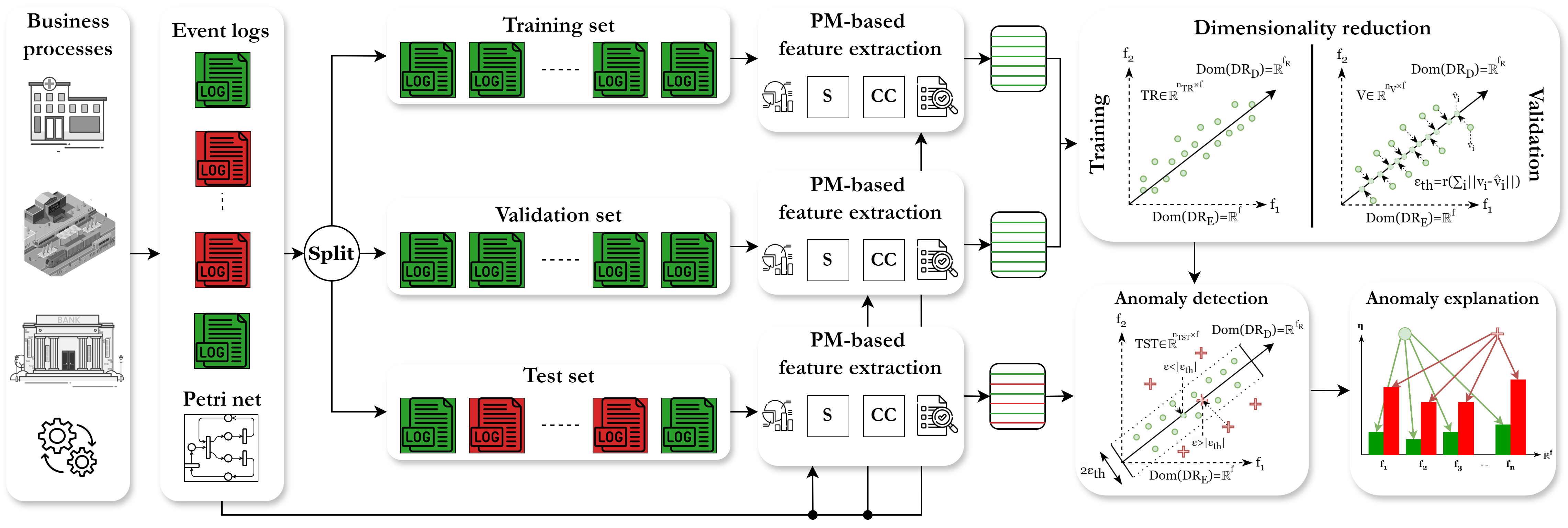}
\caption{The refined view of the proposed framework for combining \revision{process mining}-based feature extraction with dimensionality reduction for control-flow anomaly detection.}
\label{FW}
\end{figure*}

\paragraph{\revision{Explainability}}
In addition to \revision{Chandola et al's guidelines}, the framework aims to provide an explanation of the anomalies according to Li et al's \cite{li2023explainablead} definitions, which are the \textit{oracle-definition}, \textit{detection-definition}, and \textit{explanation-definition}. The \textbf{oracle-definition} is domain-specific and reflects the above-mentioned unknown, wrongly-ordered and skipped activities. For example, wrong medical treatment could lead to a doctor skipping key analyses for their patient, thus skipping activities. The remaining two types are framework-dependent and strongly connected to \revision{process mining}-based feature extraction and dimensionality reduction, which are the framework's main steps depicted in Figure \ref{FW} and described in the following.

\subsection{\revision{Process mining}-based feature extraction}
Firstly, event logs and a reference Petri net are obtained from the business process to check for deviations. On the one hand, the event logs are monitored while the business process is being executed and are labeled as normal or anomalous. On the other hand, the reference Petri net could be obtained either manually by domain experts or through the automatic generation from normal event logs, for example by using process discovery algorithms. The event logs are split into training, validation and test sets so that the normal control flow is characterized using the training set, the ability to generalize is evaluated with the validation set, and the performance is assessed with the test set, which is the only set including anomalous event logs. \revision{Process mining}-based feature extraction applies to all three sets to obtain tabular data from event data.
\begin{definition}[\revision{Process mining}-based feature extraction]
Let $\mathcal{L}\in(\mathcal{B}(\mathcal{A}^*))^n$ be an $n$-tuple of event logs and $N\in\mathcal{N}$ a Petri net. Statistical \revision{process mining}-based feature extraction $\mathcal{F}_S$ is defined as
\begin{equation}
\mathcal{F}_S:(\mathcal{B}(\mathcal{A}^*))^{n}\rightarrow\mathbb{R}^{n\times f},
\end{equation}
such that $\mathcal{F}_S(\mathcal{L})$ is an $n\times f$ matrix, where the columns are obtained directly from statistics in the event logs of $\mathcal{L}$. \revision{Conformance checking} \revision{process mining}-based feature extraction $\mathcal{F}_{CC}$ is defined as
\begin{equation}
\mathcal{F}_{CC}:(\mathcal{B}(\mathcal{A}^*))^{n}\times\mathcal{N}\rightarrow\mathbb{R}^{n\times f},
\end{equation}
such that $\mathcal{F}_{CC}(\mathcal{L},N)$ is an $n\times f$ matrix, where the columns are obtained by checking the event logs of $\mathcal{L}$ against $N$ by $CC$.
\end{definition}
Statistical \revision{process mining}-based feature extraction has the advantage of being \revision{conformance checking}-independent, as features are extracted from the statistical occurrence of certain patterns within the traces of event logs. However, the disadvantage is the absence of a reference Petri net, which allows an improved explanation of anomaly detection results. For example, Figure \ref{PETRI_NET_TRACES} shows that \revision{conformance checking} \revision{process mining}-based feature extraction implemented with our proposed approach associates the per-activity cost with model elements, flagging those transitions that cause mismatches. In the following, we review the other \revision{process mining}-based feature extraction approaches we use in Section \ref{sec:evaluation}. We denote $\mathcal{C}_{\mathcal{L}}\subseteq\mathcal{A}$ the set of activities found in the traces of the event logs of $\mathcal{L}$. Besides, recall we denote $\mathcal{C}_{\mathcal{L},N}\subseteq\mathcal{A}$ the set of activities found either in the traces of the event logs of $\mathcal{L}$ or in labeled transitions of a reference Petri net (Definition \ref{def:ab_cc}).

\paragraph{N-grams} The N-grams statistical \revision{process mining}-based feature extraction involves counting the number of times an N-tuple of activities in the traces of the event logs of $\mathcal{L}$ occurs \cite{tavares2023pmtraceencoding}. Let $\mathcal{C}_{\mathcal{L}}\times\cdots\times C_{\mathcal{L}}=(\mathcal{C}_{\mathcal{L}})^N$ be the set of N-grams and $|\mathcal{C}_{\mathcal{L}}^N|$ the total number of N-grams. $\mathcal{F}_S:(\mathcal{B}(\mathcal{\mathcal{A}^*}))^n\rightarrow\mathbb{N}^{n\times |\mathcal{C}_{\mathcal{L}}^N|}$ associates each $L\in\mathcal{L}$ with a $|\mathcal{C}_{\mathcal{L}}^N|$-tuple $(f_1,\dots, f_{|\mathcal{C}_{\mathcal{L}}^N|})$ such that $f_i\in\mathbb{N}$ counts the number of occurrences of the corresponding N-gram in $L$.

\paragraph{Directly-follows} The directly-follows statistical \revision{process mining}-based feature extraction involves counting the number of times a directly-follows relation occurs in traces of $\mathcal{L}$, where a directly-follows relation is such that, given a trace $\sigma\in L$ and activities $a_1,a_2\in\sigma$, either $a_2$ follows $a_1$ or vice versa \cite{aalst2022pmhandbook}. Let $\mathcal{DF}=(\mathcal{C}_{\mathcal{L}}\times \mathcal{C}_{\mathcal{L}})\setminus\{(c,c):c\in \mathcal{C}_{\mathcal{L}}\}$ be the set of directly-follows relations and $|\mathcal{DF}|$ the total number of directly-follows relations. $\mathcal{F}_S:(\mathcal{B}(\mathcal{\mathcal{A}^*}))^n\rightarrow\mathbb{N}^{n\times|\mathcal{DF}|}$ associates each $L\in\mathcal{L}$ with a $|\mathcal{DF}|$-tuple $(f_1,\dots,f_{|\mathcal{DF}|})$ such that $f_i\in\mathbb{N}$ counts the number of occurrences of a given directly-follows relation in $L$.

\paragraph{Token-based \revision{conformance checking}} The token-based \revision{conformance checking} \revision{process mining}-based feature extraction \cite{singh2022lapmsh, debenedictis2023dtadiiot} involves tracking the token dynamics while checking whether the traces of an event log meet the control-flow constraints of a reference Petri net. Specifically, when an activity of a trace is not allowed to execute according to the current marking of the reference Petri net, token-based \revision{conformance checking} tracks the missing ($m$) tokens. In like manner, when the last activity of a trace executes, token-based \revision{conformance checking} tracks the remaining ($r$) tokens in the places of the reference Petri net. Finally, token-based \revision{conformance checking} tracks the consumed ($c$) and produced ($r$) tokens, and the total number of activities executed. Formally, $\mathcal{F}_{CC}:(\mathcal{B}(\mathcal{A}^*))^n\times\mathcal{N}\rightarrow \mathbb{R}^{n\times 3+|\mathcal{C}_{\mathcal{L},N}|}$ associates each $L\in\mathcal{L}$ with a $(3+|\mathcal{C}_{\mathcal{L},N}|)$-tuple $(F_{L,N}, f_m, f_r, f_{\mathcal{C}_{\mathcal{L},N},1},\dots,f_{\mathcal{C}_{\mathcal{L},N},|\mathcal{C}_{\mathcal{L},N}|})$ such that $F_{L,N}=\frac{1}{|L|}\sum_{\sigma\in L}\frac{1}{2}(1-\frac{m}{c})+\frac{1}{2}(1-\frac{r}{p})$ is the fitness obtained by token-based \revision{conformance checking} \cite{aalst2022pmhandbook}, $f_m, f_r\in\mathbb{N}$ count the number of missing and remaining tokens, and $f_{\mathcal{C}_{\mathcal{L},N},i}$ counts the number of times a given activity $c_i\in\mathcal{C}_{\mathcal{L},N}$ executed.

\subsection{Dimensionality reduction}
Regardless of the type of \revision{process mining}-based feature extraction used, the resulting tabular data are managed by a dimensionality reduction technique that reconstructs the data by encoding and decoding functions. This allows control-flow anomaly detection by the reconstruction-based approach, which has shown satisfying performance on account of the ability of dimensionality reduction to handle high-dimensional and possibly non-linear feature spaces.
\begin{definition}[Dimensionality reduction technique]
Let $T\in\mathbb{R}^{n\times f}$ be a matrix with $n$ rows and $f$ columns. A dimensionality reduction technique $\mathcal{DR}$ is a pair of functions defined as follows:
\begin{equation}
\begin{cases}
\mathcal{DR}_E:\mathbb{R}^{f}\rightarrow\mathbb{R}^{f_R}\\
\mathcal{DR}_D:\mathbb{R}^{f}\rightarrow\mathbb{R}^{f}
\end{cases}\quad f_R<f,
\end{equation}
where $\mathcal{DR}_E$ is the encoding function and $\mathcal{DR}_D$ the decoding function. $\hat{t}_i=\mathcal{DR}_D(t_i)$ is the reconstruction of the $i$-th row of $T$ by the decoding function of the technique after obtaining the encoding model upon the application of the encoding function. Let us define $\mathcal{E}:\mathbb{R}^f\rightarrow \mathbb{R}$ the function that calculates the reconstruction error. We denote $\mathcal{E}_{t_i}=\mathcal{E}(t_i)= ||t_i-\hat{t}_i||$ the reconstruction error of $t_i$.
\end{definition}

The dimensionality reduction phase is two-stepped. These are the training and validation steps. Firstly, the encoding part of a dimensionality reduction technique $\mathcal{DR}_E$ is used to project the training tabular data $TR\in\mathbb{R}^{n_{TR}\times f}$ onto a new lower-dimensional coordinate system with domain $\mathbb{R}^{f_R}$. This coordinate system is used to project validation tabular data $V\in\mathbb{R}^{n_V\times f}$ onto it. After this projection, each validation sample is reconstructed in the original space with the decoding part of the dimensionality reduction technique $\mathcal{DR}_D$. The process repeats with different configurations of the dimensionality reduction technique until the reconstruction error on the validation set is minimized. In the following, we review the dimensionality reduction techniques we use in Section \ref{sec:evaluation}. We denote $T\in\mathbb{R}^{n\times f}$ a matrix obtained by \revision{process mining}-based feature extraction with centered columns, i.e., the mean of each column is 0 and their standard deviation is 1, $t\in \mathbb{R}^f$ a row vector of $T$, and $X'$ ($x'$) the transpose of a generic matrix (row vector) $X$ ($x$).

\paragraph{Principal Component Analysis} The Principal Component Analysis (PCA) is a widespread linear approach for dimensionality reduction \cite{abdi2010pca}. The PCA projects $T$ onto a new coordinate system by eigendecomposition of the $T$'s covariance matrix: $\frac{1}{n}T'T=U\Delta U'$, where $U\in\mathbb{R}^{f\times f}$ is the matrix whose columns are eigenvectors and $\Delta\in\mathbb{R}^{f\times f}$ is a diagonal matrix containing the eigenvalues corresponding to the eigenvectors. The columns of $U$ represent the so-called principal components, whose linear combination with the rows of $T$ results in the projection of $T$ onto the new coordinate system. The PCA is such that the first few principal components concentrate most of the explained variance. Hence, on the one hand, the encoding part of the PCA can be formulated as $\mathcal{DR}_E(t)=tU_P$, where $U_P\in\mathbb{R}^{f\times f_R}$ is $U$ with the first $f_R$ columns retained. On the other hand, the decoding part of the PCA can be formulated as $\mathcal{DR}_D(t)=\mathcal{DR}_E(t)U_P'$.

\paragraph{Sparse Principal Component Analysis} The Sparse PCA (SPCA) extends the PCA by leaving out some features of the input space when building the new coordinate system to include only the significant ones \cite{zou2006sparsepca}. This is enforced by converting the PCA formulation as a sparse regression problem with a regression regularization parameter and a sparsity coefficient, leading to the sparse matrix $U_S\in\mathbb{R}^{f\times f}$ and the reduced version $U_{S,P}\in\mathbb{R}^{f\times f_R}$. The encoding and decoding parts of the SPCA are the same as the PCA.

\paragraph{Kernel Principal Component Analysis} The Kernel PCA (KPCA) extends the PCA by finding a high-dimensional hyperplane where $T$ can be linearly separated using the so-called kernel trick \cite{nguyen2010kpcafd}. Firstly, each row of $T$ is projected onto a higher-dimensional hyperplane with domain $\mathbb{R}^{f_e}, f_e>f$ by a mapping $\phi: \mathbb{R}^f \rightarrow \mathbb{R}^{f_e}$. The covariance matrix of the transformed (centered) data is as follows: $C=\frac{1}{n}\sum_{i=1}^n\phi(t_i)'\phi(t_i)\in\mathbb{R}^{f_e\times f_e}$, where $t_i\in\mathbb{R}^f$ is a row of $T$. Let us consider the kernel function $k(t_i,t_j)=\phi(t_i)\phi(t_j)'\in\mathbb{R}$ and the kernel matrix $K\in\mathbb{R}^{n\times n}$ such that $K_{ij}=k(t_i,t_j)$. The eigenvalue equation $\lambda V=CV$, where $\lambda\in\mathbb{R}$ and $V\in \mathbb{R}^{f_e}$, transforms to $K\nu = n\lambda\nu$, where $\nu\in \mathbb{R}^n$ is an eigenvector of $K$. The eigenvectors represent the kernel principal components, leading to the matrix $U_e\in\mathbb{R}^{n\times n}$ and the reduced version $U_{e,P}\in\mathbb{R}^{n\times f_R}$. The encoding part of the KPCA can be formulated as $\mathcal{DR}_E(t)=\kappa(t)U_{e,P},\kappa(t)=(\phi(t,t_1),\dots,\phi(t,t_n))\in\mathbb{R}^{n}$. The decoding part involves ridge regression, which requires solving the following minimization problem: $\min_{\beta}\{||U_{e,P}\beta-T||^2+\gamma||\beta||^2\}$, where $||\cdot||^2$ indicates the squared norm, $\gamma\in\mathbb{R}$ is a regularization term, and $\beta\in\mathbb{R}^{f_e\times f}$ the matrix of regression components. Finally, the decoding part of the KPCA can be formulated as $\mathcal{DR}_{D}(t)=\kappa(t)\beta$.

\paragraph{Autoencoder} The autoencoder is a widespread feed-forward neural network approach for dimensionality reduction which accounts for non-linear dependencies in $T$'s features through the non-linear transformations of the network's neurons \cite{sakurada2014adeutoencoders}. The simplest autoencoder structure is a symmetric feed-forward network with three fully-connected hidden layers, namely the encoder, code and decoder layers. The code layer implements $\mathcal{DR}_E(t)=W_C\sigma_E(W_Et+b_E)+b_C$, where $b_E\in\mathbb{R}^{f_{E,D}}$ and $b_C\in\mathbb{R}^{f_R}$ are the encoder and code bias vectors, $W_E\in\mathbb{R}^{f_{E,D}\times f}$ and $W_C\in\mathbb{R}^{f_R \times f_{E,D}}$ are the encoder and code weight matrices, and $\sigma_E$ the activation function applied by the encoder layer's neurons. The output provides the reconstructed data, implementing the decoding part $\mathcal{DR}_D(t)=W_O\sigma_D(W_D\mathcal{DR}_E(t)+b_D)+b_O$, where $b_D\in\mathbb{R}^{f_{E,D}}$ and $b_O\in\mathbb{R}^{f}$ are the decoder and output bias vectors, $W_D\in\mathbb{R}^{f_R\times f_{E,D}}$ and $W_O\in\mathbb{R}^{f \times f_{E,D}}$ are the decoder and output weight matrices, and $\sigma_D$ the activation function applied by the decoder layer's neurons.

Table \ref{DR_TECHNIQUES} briefly describes the techniques above, mentioning the approach followed, the complexity and type of the encoding and decoding parts, and the framework formulation. The choice of the specific technique depends on the computational resources available and the data at hand. For example, on the one hand, the PCA can be preferred to other methods if the feature space of data does not exhibit non-linear relationships and/or there are limited computational resources available. On the other hand, if computational resources are not constrained and there are loads of data with non-linear relationships in the feature space, the autoencoder could provide better results. Finally, it is worth mentioning that there are other dimensionality reduction techniques, such as the linear discriminant analysis \cite{tharwat2017lda} and t-distributed stochastic neighbor embedding \cite{belkina2019tsne}, as well as further extensions of the ones we presented, such as the robust PCA \cite{bouwmans2014rpca} and variational autoencoder \cite{kingma2014vae}.
\begin{table}[!t]
\centering
\caption{A set of dimensionality reduction techniques with their framework formulation.}
\label{DR_TECHNIQUES}
\resizebox{\columnwidth}{!}{%
\begin{tabular}{ccccc}
\toprule
\textbf{Technique} & \textbf{Approach}                                                                                        & \textbf{Complexity} & \textbf{Type} & \textbf{Formulation}                                                                                                                                  \\ \midrule
\textbf{PCA}       & Eigendecomposition                                                                                       & Low                 & Linear        & \begin{tabular}[c]{@{}c@{}}$\mathcal{DR}_E(t)=tU_P$\\ $\mathcal{DR}_D(t)=\mathcal{DR}_E(t)U_P'$\end{tabular}                                          \\
\textbf{SPCA}      & Sparse regression                                                                                        & Medium              & Linear        & \begin{tabular}[c]{@{}c@{}}$\mathcal{DR}_E(t)=tU_{S,P}$\\ $\mathcal{DR}_D(t)=\mathcal{DR}_E(t)U_{S,P}'$\end{tabular}                                  \\
\textbf{KPCA}      & \begin{tabular}[c]{@{}c@{}}Kernel trick,\\ eigendecomposition,\\ kernel ridge regression\end{tabular} & Medium                & Non-linear    & \begin{tabular}[c]{@{}c@{}}$\mathcal{DR}_E(t)=\kappa(t)U_{e,P}$\\ $\mathcal{DR}_D(t)=\kappa(t)\beta$\end{tabular}                                     \\
\textbf{Autoencoder}        & Neural network                                                                              & High                & Non-linear    & \begin{tabular}[c]{@{}c@{}}$\mathcal{DR}_E(t)=W_C\sigma_E(W_Et+b_E)+b_C$\\ $\mathcal{DR}_D(t)=W_O\sigma_D(W_D\mathcal{DR}_E(t)+b_D)+b_O$\end{tabular} \\ \bottomrule
\end{tabular}%
}
\end{table}

\subsection{Anomaly detection and explanation}
A threshold on the reconstruction error $\mathcal{E}_{th}$ can be calculated for classifying test data during anomaly detection. Let $\mathcal{E}_{v_i}=||v_i-\hat{v}_i||,i\in\{1,\dots,n_V\}$ be the reconstruction errors of rows $v_i\in\mathbb{R}^{f}$ of validation tabular data $V\in\mathbb{R}^{n_V\times f}$. Let $\sum_i||v_i-\hat{v}_i||$ be the sum of such errors. $\mathcal{E}_{th}$ is calculated by a function $r$ over the sum of reconstruction errors, i.e., $\mathcal{E}_{th}=r(\sum_i||v_i-\hat{v}_i||)$. For example, $r$ can be the mean squared error: $\mathcal{E}_{th}=\frac{\sum_i ||v_i-\hat{v}_i||^2}{n_V}$.
\begin{definition}[Anomaly detection]
\label{def:ad}
Let $t\in\mathbb{R}^{f}$ be a test sample, $\mathcal{DR}_D$ the decoding part of a dimensionality reduction technique tailored to a training set, $\hat{t}=\mathcal{DR}_D(t)$ the reconstruction of $t$ using $\mathcal{DR}_D$, $\mathcal{E}_t=||t-\hat{t}||$ the reconstruction error of $t$, and $\mathcal{E}_{th}$ a threshold to the reconstruction error obtained with a validation set. The classification of $t$ is \emph{normal} if $\mathcal{E}_{t}<\mathcal{E}_{th}$, \emph{anomalous} otherwise.
\end{definition}

The combination of a \revision{process mining}-based feature extraction approach and a dimensionality reduction technique forms a \textit{framework technique}. The \textbf{detection-definition} of framework techniques is linked to Definition \ref{def:ad}: the framework techniques identify as anomalous those event logs whose \revision{process mining}-based feature extraction leads to a reconstruction error higher than the threshold. However, while this definition outlines \textit{what is an anomaly}, it does not provide insight into \textit{why} the event log led to an anomalous reconstruction error. There are several anomaly explanation techniques available to provide an \textbf{explanation-definition}. In this work, we focus on additive feature-based explanation, which explains the output of an anomaly detection technique by superposition of the impact of the features of input data \cite{lundberg2017unifiedapproach}.

\begin{definition}[Anomaly explanation]
\label{def:ad_exp}
Let $t\in\mathbb{R}^{f}$ be a test sample, $\mathcal{DR}_D$ the decoding part of a dimensionality reduction technique tailored to a training set, $\mathcal{DR}_D(t)=\hat{t}$ the reconstruction of $t$ using $\mathcal{DR}_D$, and $\mathcal{E}$ the function that calculates the reconstruction error. The additive explanation of $\mathcal{E}_t$ is $\tilde{\mathcal{E}}_t=\theta+\sum_{i=1}^f\eta_i\mathcal{E}_t,\theta,\eta_{i}\in\mathbb{R}$. $\eta=(\eta_1,\dots,\eta_f)\in\mathbb{R}^f$ is the \emph{anomaly explanation}, where $\eta_i$ denotes the effect of feature $f_i$ on $\mathcal{E}_t$.
\end{definition}

Several methods such as Local Interpretable Model-agnostic Explanations, DeepLift, and Classic Shapley Value Estimation can be used to calculate $\eta$. These methods are unified under a framework that calculates the so-called SHapley Additive exPlanations (SHAP) values \cite{lundberg2017unifiedapproach}. These values have been used to explain the results of several dimensionality reduction approaches to anomaly detection \cite{takeishi2019pcashapexplanation,antwarg2021aeshapexplanation}. In conclusion, we would like to remark that not only does this explanation-definition depend on the dimensionality reduction technique but also on \revision{process mining}-based feature extraction. For example, SHAP values linked to the features of our proposed \revision{process mining}-based feature extraction approach directly connect to the elements of the reference Petri net. This connection gives additional model-based insight into the reason why test data are classified as anomalous.

\section{Experiments}
\label{sec:evaluation}
In this section, we first describe the datasets used. Secondly, we detail the framework techniques and baseline \revision{conformance checking}-based techniques with which we experiment. Finally, the goals of the experiments, the performance metrics used, and the results of the experiments are reported and discussed.

\subsection{Datasets}
The datasets include well-known benchmarks for anomaly detection in event logs (PDC 2020/2021), simulation related to a railways case study (ERTMS), and real-world data from a healthcare case study (COVAS).

\paragraph{PDC 2020/2021}
The PDC datasets\footnote{\url{https://www.tf-pm.org/competitions-awards/discovery-contest}} are generated by simulation of a variety of Petri nets. For each PDC dataset, the different Petri nets are obtained by tuning several characteristics of a base Petri net, introducing more complex control-flow patterns such as non-local dependencies between activities. PDC 2020 contains two event log sets named \texttt{ground\_truth} and \texttt{training}. Each event log of the \texttt{training} set is generated by randomly walking one of the Petri net variants 1000 times, thus producing 1000 traces. In addition to simulating the Petri net variant, the training set contains an event log corresponding to a Petri net variant with noise introduced in the traces. For example, given $N$ the Petri net variant and $L_N$ the simulated event log of the \texttt{training} set, the authors produce the noisy event log $\tilde{L}_N$.
Regarding the \texttt{ground\_truth} set, there are as many event logs of 1000 traces as Petri net variants. However, the authors did not mention whether these event logs were obtained by randomly walking the corresponding Petri net variant. Instead, they label each trace as either \texttt{positive} or \texttt{negative} to allow practitioners to test their algorithms. In our experimentation, we organize the dataset as follows. We take the \texttt{training} event logs of two Petri net variants and the corresponding \texttt{ground\_truth} event logs. Then, we consider normal one of the two variants, and anomalous the other. Specifically, all the traces of the \texttt{training} event logs of the normal variant are considered normal, and \texttt{positive} traces of the \texttt{ground\_truth} event log are also considered normal. Conversely, all traces of the \texttt{training} event logs of the anomalous variant are considered anomalous, and \texttt{positive} traces of the \texttt{ground\_truth} event log are also considered anomalous because they fit the anomalous variant. The same process is applied to the PDC 2021 dataset. Finally, we consider the base Petri net in both datasets as the input Petri net for \revision{conformance checking} \revision{process mining}-based feature extraction.

\paragraph{ERTMS}
The European Rail Traffic Management System (ERTMS)\footnote{\url{https://www.ertms.net/}} is a standard that regulates European railways for the smooth and safe operation of trains across Europe. ERTMS prescribes textual requirements for several procedures and provides informal activity and state diagrams to facilitate the implementation of these procedures. In this work, we consider the RBC/RBC Handover procedure, which regulates the process that trains must follow when their journey involves changing their supervision from one RBC to another, where the RBC is an entity whose supervision scope involves a limited area of on-track equipment and trains. This procedure was considered in \cite{debenedictis2023dtadiiot} and used as proof of concept for the experimentation of anomaly detection in simulated traces generated by a BPMN model of the RBC/RBC Handover. We replicate the anomaly injection process proposed by the authors as follows. We consider the resources associated with the activities of the BPMN model of the RBC/RBC Handover and randomly walk the model to generate 2000 traces. Then, we split these 2000 traces into two normal and anomalous event logs of 1000 traces each. Each trace of both event logs is injected with anomalies. This injection: randomly selects a resource among those involved in RBC/RBC Handover, considers the activities associated with this resource, and proceeds to introduce missed, duplicated, or wrongly-ordered activities in the trace uniformly according to a given probability. This probability depends on the nature of the trace. In our simulations, we set the probability to 5\% for each anomaly type in normal traces and 25\% for each anomaly type in anomalous traces. This means that, given a normal trace and assuming the introduction of each anomaly is independent, there is $1-0.95^3\approx15\%$ probability of having at least one of the anomalies, whereas, given an anomalous trace and assuming the same, there is $1-0.75^3\approx57\%$ probability of having at least one of the anomalies.

\paragraph{COVAS}
The COVID-19 Aachen Study (COVAS) involved monitoring the treatment process of COVID-19 patients from March 2020 to June 2021 \cite{pegoraro2021covas,benevento2022covas}. The data are preprocessed by cleaning incomplete traces, using abstractions for some activities of negligible significance, and filtering infrequent variants. Furthermore, the data are split into three event logs, according to three different time frames matching the first, second and third infection waves. This split is backed up by literature, as there have been significant differences in COVID-19 patients' clinical course and treatment among these three infection waves. The three COVAS event logs have 106, 59, and 22 traces. The traces are labeled following the viewpoint of the first infection wave. Hence, the first-wave event log contains normal traces, whereas the other second- and third-wave event logs are anomalous. To achieve a dataset whose dimension is comparable to the others, we augment the data by random re-sampling of traces of the same kind. Finally, we consider a Petri net related to the first infection wave and discovered by authors in \cite{benevento2022covas} as the input Petri net for \revision{conformance checking} \revision{process mining}-based feature extraction.

For all datasets, the normal and anomalous traces are grouped into event logs of 5 traces each, resulting in two sets of normal and anomalous event logs. All anomalous event logs are retained in the test set, whereas the normal event logs are split as follows. 25\% of normal event logs are held out and placed into the test set and the remainder is used as the training set. Next, 20\% of the training set is held out and placed into the validation set. Table \ref{DATASETS_INFO} collects the properties of the datasets. These properties outline whether the target application is real or abstract, the nature is synthetic or real-world, and the labeling is synthetic or done by domain experts. For example, let us consider the ERTMS dataset. It is a real application because it involves a railway standard, its nature is synthetic because the data is obtained by simulating the model, and the labels are synthetic since they depend on the probabilistic injection of anomalies. Furthermore, we report the number of normal and anomalous traces, and the resulting training, validation and test event logs obtained according to the percentages above. 
\begin{table}[!t]
\centering
\caption{The properties of the datasets for the experiments.}
\label{DATASETS_INFO}
\resizebox{\columnwidth}{!}{%
\begin{tabular}{lllll}
\toprule
\textbf{Property}              & \textbf{COVAS}    & \textbf{ERTMS} & \textbf{PDC 2020} & \textbf{PDC 2021} \\ \midrule
\textbf{Application}           & Real              & Real           & Abstract          & Abstract          \\
\textbf{Nature}                & Real-world        & Synthetic      & Synthetic         & Synthetic         \\
\textbf{Labeling}              & By domain experts & Synthetic      & Synthetic         & Synthetic         \\
\textbf{Normal traces}         & 1000 (augmented)             & 1000           & 3020              & 4250              \\
\textbf{Anomalous traces}      & 1000 (augmented)             & 1000           & 2496              & 2125              \\
\textbf{Training event logs}   & 120               & 120            & 363               & 511               \\
\textbf{Validation event logs} & 30                & 30             & 90                & 127               \\
\textbf{Test event logs}       & 250               & 250            & 650               & 637               \\ \bottomrule
\end{tabular}%
}
\end{table}
\subsection{Techniques}
\subsubsection{Baselines}
In this work, we consider the \revision{conformance checking}-based techniques that rely on fitness thresholding as the baselines \cite{bezerra2009pmad, bezerra2013adlogspais, myers2018icsadpm, pecchia2020applicationfailuresanalysispm}. As mentioned, Figure \ref{CC_UNBALANCING} highlights that setting a threshold for the fitness measure is challenging since the fitness values are spread inconsistently, even though the normal event log and the reference Petri net are, respectively, the set of first-wave traces and Petri net from the COVAS case study. This is a critical issue, especially because the reference Petri net was obtained after laborious pre-processing, hence it should be of reasonable quality. Regardless, \revision{conformance checking}-based techniques that rely on fitness thresholding are worth comparing due to their explainable nature. In fact, the fitness measure comes from a clear-cut model-based definition (e.g., Definition \ref{def:ab_fitness}), which provides a straightforward explanation of why an event log is classified as anomalous.

We implement two baseline \revision{conformance checking}-based techniques. Firstly, token-based/alignment-based \revision{conformance checking} is applied to all event logs in the validation set against the reference Petri net to obtain the fitness measures for all the validation event logs. Secondly, a threshold is automatically assigned using the minimum of all the fitness measures. This is because, given that every event log in the validation set is normal, the best generalization should be achieved by considering the least fitting event log. We name AB\_CC\_B and TB\_CC\_B the baselines employing token-based and alignment-based \revision{conformance checking}, respectively.

\subsubsection{Framework}
The framework allows implementing a wide range of \revision{conformance checking}-based and \revision{conformance checking}-independent control-flow anomaly detection techniques by combining a \revision{process mining}-based feature extraction approach with a dimensionality reduction technique. This combination allows assessing the impact of diverse \revision{process mining}-based feature extraction approaches on the results \cite{ko2023adsystematicreview, tavares2023pmtraceencoding} and replicating the reconstruction-based approach for control-flow anomaly detection \cite{nolle2018processadautoencoders, vijayakamal2020bpaead, elaziz2023drlbpad, chinnaiah2024deepaead, kan2024aebasedelad}.

We implement 16 framework techniques for control-flow anomaly detection. These techniques combine the proposed (Section \ref{sec:abccfe}) and reviewed (Section \ref{sec:framework}) \revision{process mining}-based feature extraction approaches, namely the alignment-based \revision{conformance checking} (AB\_CC), token-based \revision{conformance checking} (TB\_CC), N-grams (NG) and directly-follows (DF) approaches, and the reviewed (Section \ref{sec:framework}) dimensionality reduction techniques, namely the PCA, SPCA, KPCA and autoencoder (AE). Table \ref{DR_TECHNIQUES_CONFIGURATIONS} provides an overview of the parameters used to configure the dimensionality reduction techniques during the validation step of the framework, which performs an exhaustive search of all the parameter values to minimize the reconstruction error. Each framework technique is labeled by a pair of values (PF, DR), where PF $\in$ \{AB\_CC, TB\_CC, NG, DF\} indicates the \revision{process mining}-based feature extraction approach and DR $\in$ \{PCA, KPCA, SPCA, AE\} the dimensionality reduction technique.

\begin{table*}[!t]
\centering
\caption{Parameters used to configure the dimensionality reduction techniques during the validation step of the framework. N/A = Not Applicable.}
\label{DR_TECHNIQUES_CONFIGURATIONS}
\resizebox{\textwidth}{!}{%
\begin{tabular}{ccccccc}
\toprule
\textbf{$\boldsymbol{\mathcal{DR}}$} & \textbf{Parameter \#1}                                          & \textbf{Parameter \#2}                                                                                  & \textbf{Parameter \#3}                                                     & \textbf{Parameter \#4}                                                                           & \textbf{Parameter \#5}                                                     & \textbf{Parameter \#6}                                                 \\ \midrule
\textbf{PCA}            & \begin{tabular}[c]{@{}c@{}}$f_R$\\ \{2, 4, 8, 16\}\end{tabular} & N/A                                                                                                     & N/A                                                                        & N/A                                                                                              & N/A                                                                        & N/A                                                                    \\
\textbf{SPCA}           & \begin{tabular}[c]{@{}c@{}}$f_R$\\ \{2, 4, 8, 16\}\end{tabular} & \begin{tabular}[c]{@{}c@{}}Regression regularization\\ \{0.01, 0.1, 0.25, 0.5, 0.75, 1.0\}\end{tabular} & \begin{tabular}[c]{@{}c@{}}Sparsity coefficient\\ \{0.1, 0.5, 1, 2, 3\}\end{tabular}   & N/A                                                                                              & N/A                                                                        & N/A                                                                    \\
\textbf{KPCA}           & \begin{tabular}[c]{@{}c@{}}$f_R$\\ \{2, 4, 8, 16\}\end{tabular} & \begin{tabular}[c]{@{}c@{}}Regression regularization\\ \{0.01, 0.1, 0.25, 0.5, 0.75, 1.0\}\end{tabular} & \begin{tabular}[c]{@{}c@{}}Kernel\\ \{poly, rbf, sigmoid\}\end{tabular}    & \begin{tabular}[c]{@{}c@{}}Kernel coefficient\\ \{0.01, 0.05, 0.1\}\end{tabular} & \begin{tabular}[c]{@{}c@{}}Polynomial degree\\ \{3, 4, 5, 6\}\end{tabular} & N/A                                                                    \\
\textbf{Autoencoder}             & \begin{tabular}[c]{@{}c@{}}$f_R$\\ \{2, 4, 8, 16\}\end{tabular} & \begin{tabular}[c]{@{}c@{}}Hidden neurons\\ \{32, 64, 128, 256\}\end{tabular}                           & \begin{tabular}[c]{@{}c@{}}Optimizer\\ \{adam, rmsprop, SGD\}\end{tabular} & \begin{tabular}[c]{@{}c@{}}Batch size\\ \{8, 16, 32, 64\}\end{tabular}                           & \begin{tabular}[c]{@{}c@{}}Epochs\\ \{100, 250, 500\}\end{tabular}         & \begin{tabular}[c]{@{}c@{}}Activation function\\ \{ReLu\}\end{tabular} \\ \bottomrule
\end{tabular}%
}
\end{table*}

\subsection{Evaluation}
The experiments are designed to evaluate two aspects of our research, which are the following: 1) compare to the baseline \revision{conformance checking}-based techniques the detection effectiveness and explainability of the framework techniques using our proposed \revision{process mining}-based feature extraction approach; and 2) compare the detection effectiveness of \revision{conformance checking}-based and \revision{conformance checking}-independent framework techniques and evaluate the factors that impact their performance. The performance of the techniques is evaluated by their accuracy ($A$), recall ($R$), precision ($P$) and F1-Score ($F1$).
\begin{definition}[Accuracy, recall, precision and F1-score]
Let $\mathcal{L}\in(\mathcal{B}(\mathcal{A}^*))^n$ be an $n$-tuple of labeled test event logs. Let positive samples be normal event logs and negative samples be anomalous event logs. Finally, let $TP$, $TN$, $FP$ and $FN$ be, respectively, the true positives, true negatives, false positives, and false negatives. Accuracy $A$, recall $R$, precision $P$, and F1-Score $F1$ are defined as follows:
\begin{equation*}
A=\frac{TP+TN}{TP+TN+FP+FN};\,R=\frac{TP}{TP+FN};
\end{equation*}
\begin{equation}
    P=\frac{TP}{TP+FP};\,F1=\frac{2RP}{R+P}.
\end{equation}
\end{definition}

The software for the experiments is implemented using Python and was run on a Windows 10 machine with an Intel® Core™ i9-11900K CPU @ 3.50GHz and 32GB of RAM memory. The software uses libraries for machine learning and \revision{process mining}, such as \texttt{scikit-learn}, \texttt{tensorflow} and \texttt{pm4py}. In addition, it includes a set of DOS batch scripts to automatically replicate the experiments carried out in this work, available online on GitHub\footnote{\url{https://github.com/francescovitale/pm_dr_framework}}.

\subsubsection{Baselines comparison}
In this part, we compare the detection effectiveness of the framework techniques using our proposed \revision{process mining}-based feature extraction approach, namely (AB\_CC, PCA), (AB\_CC, SPCA), (AB\_CC, KPCA) and (AB\_CC, AE), with the baselines. Each technique is applied to each dataset 5 times to account for statistical fluctuations due to the random split of training, validation and test sets.

\paragraph{Results} Table \ref{BASELINES_COMPARISON} reports the mean $A$, $R$, $P$ and $F1$ measures and their standard deviation in subscripts for each technique and dataset. Except for AB\_CC\_B's $R$ measure for the COVAS dataset (89.1\%), the \revision{conformance checking}-based framework techniques using our proposed \revision{process mining}-based feature extraction approach outperform both baselines, in some cases by a very large degree. For the PDC 2020 and 2021 datasets, the (AB\_CC, AE) technique achieves 97.3\% and 87.2\% $F1$, whereas the AB\_CC\_B and TB\_CC\_B baselines achieve 36.1\% and 19.4\% $F1$ for the PDC 2020 dataset, and 33.4\% and 56.6\% $F1$ for the PDC 2021 dataset. The performance gap is smaller for the ERTMS and COVAS datasets, for which (AB\_CC, AE) and (AB\_CC, KPCA) achieve 85.2\% and 88.5\% $F1$, whereas the AB\_CC\_B and TB\_CC\_B baselines both achieve 78.8\% $F1$ for the ERTMS dataset, and 79.1\% and 18.4\% $F1$ for the COVAS dataset.

\begin{table*}[!t]
\centering
\caption{The mean $A$, $R$, $P$ and $F1$ measures and their standard deviation in subscripts for each framework/baseline technique and dataset. Measures in bold indicate the highest figure per dataset.}
\label{BASELINES_COMPARISON}
\resizebox{\textwidth}{!}{%
\begin{tabular}{llllllllllllllllllllll}
\toprule
\multicolumn{2}{l}{\textbf{}}          &  & \multicolumn{4}{l}{\textbf{PDC 2020}}                                                         &  & \multicolumn{4}{l}{\textbf{PDC 2021}}                                                         &  & \multicolumn{4}{l}{\textbf{ERTMS}}                                                            &  & \multicolumn{4}{l}{\textbf{COVAS}}                                                            \\ \cmidrule{4-7} \cmidrule{9-12} \cmidrule{14-17} \cmidrule{19-22} 
\multicolumn{2}{l}{\textbf{Technique}} &  & \textbf{A(\%)}        & \textbf{R(\%)}        & \textbf{P(\%)}        & \textbf{F1(\%)}       &  & \textbf{A(\%)}        & \textbf{R(\%)}        & \textbf{P(\%)}        & \textbf{F1(\%)}       &  & \textbf{A(\%)}        & \textbf{R(\%)}        & \textbf{P(\%)}        & \textbf{F1(\%)}       &  & \textbf{A(\%)}        & \textbf{R(\%)}        & \textbf{P(\%)}        & \textbf{F1(\%)}       \\ \midrule
\multicolumn{2}{l}{(AB\_CC, PCA)}      &  & 96.2$_{1.0}$          & 92.7$_{1.2}$          & 96.5$_{2.2}$          & 94.4$_{1.4}$          &  & 86.0$_{2.4}$          & 83.5$_{2.4}$          & 84.8$_{3.0}$          & 84.0$_{2.6}$          &  & 87.6$_{1.3}$          & 76.8$_{4.6}$          & 82.3$_{1.6}$          & 78.8$_{3.4}$          &  & 85.1$_{6.4}$          & 77.5$_{5.8}$          & 78.5$_{9.9}$          & 77.5$_{7.7}$          \\
\multicolumn{2}{l}{(AB\_CC, SPCA)}     &  & 96.8$_{1.0}$          & 93.8$_{1.4}$          & 97.1$_{2.0}$          & 95.3$_{1.4}$          &  & 86.3$_{2.9}$          & 83.1$_{2.6}$          & 85.8$_{3.8}$          & 84.1$_{3.1}$          &  & 88.2$_{2.3}$          & 78.8$_{5.5}$          & 82.5$_{3.4}$          & 80.3$_{4.7}$          &  & 91.0$_{4.8}$          & 87.6$_{7.2}$          & 85.8$_{7.5}$          & 86.5$_{7.0}$          \\
\multicolumn{2}{l}{(AB\_CC, KPCA)}     &  & 96.4$_{0.7}$          & 92.1$_{1.5}$          & 97.8$_{0.4}$          & 94.6$_{1.1}$          &  & 86.6$_{1.4}$          & 83.5$_{2.3}$          & 85.9$_{1.6}$          & 84.4$_{1.9}$          &  & 89.3$_{1.7}$          & \textbf{87.8$_{2.3}$} & 83.1$_{3.4}$          & 84.7$_{1.7}$          &  & \textbf{92.8$_{1.0}$} & 87.9$_{3.2}$          & \textbf{89.5$_{1.1}$} & \textbf{88.5$_{2.0}$} \\
\multicolumn{2}{l}{(AB\_CC, AE)}       &  & \textbf{98.1$_{0.3}$} & \textbf{96.6$_{1.3}$} & \textbf{98.1$_{0.5}$} & \textbf{97.3$_{0.5}$} &  & \textbf{88.9$_{1.1}$} & \textbf{86.2$_{1.3}$} & \textbf{88.4$_{1.4}$} & \textbf{87.2$_{1.3}$} &  & \textbf{91.2$_{1.9}$} & 83.3$_{5.8}$          & \textbf{88.3$_{1.6}$} & \textbf{85.2$_{4.1}$} &  & 91.4$_{3.1}$          & 85.6$_{6.5}$          & 87.0$_{5.0}$          & 86.2$_{5.7}$          \\ \midrule
\multicolumn{2}{l}{AB\_CC\_B}          &  & 36.6$_{4.3}$          & 58.6$_{2.2}$          & 62.3$_{0.5}$          & 36.1$_{4.0}$          &  & 38.7$_{4.6}$          & 53.8$_{3.2}$          & 65.9$_{1.9}$          & 33.4$_{6.7}$          &  & 82.3$_{10.0}$         & 87.6$_{5.6}$          & 78.2$_{8.7}$          & 78.8$_{10.3}$         &  & 82.8$_{4.9}$          & \textbf{89.1$_{2.7}$} & 77.3$_{4.4}$          & 79.1$_{5.2}$          \\
\multicolumn{2}{l}{TB\_CC\_B}          &  & 23.1$_{0.2}$          & 49.9$_{0.6}$          & 54.7$_{7.6}$          & 19.4$_{0.5}$          &  & 57.0$_{2.4}$          & 67.6$_{1.8}$          & 71.4$_{0.7}$          & 56.6$_{2.7}$          &  & 81.9$_{11.9}$         & 87.7$_{6.5}$          & 78.7$_{9.5}$          & 78.8$_{12.0}$         &  & 21.2$_{0.0}$          & 50.4$_{0.0}$          & 56.9$_{4.0}$          & 18.4$_{0.0}$          \\ \bottomrule
\end{tabular}%
}
\end{table*}
\begin{figure*}[!t]
\centering
\includegraphics[width=\textwidth]{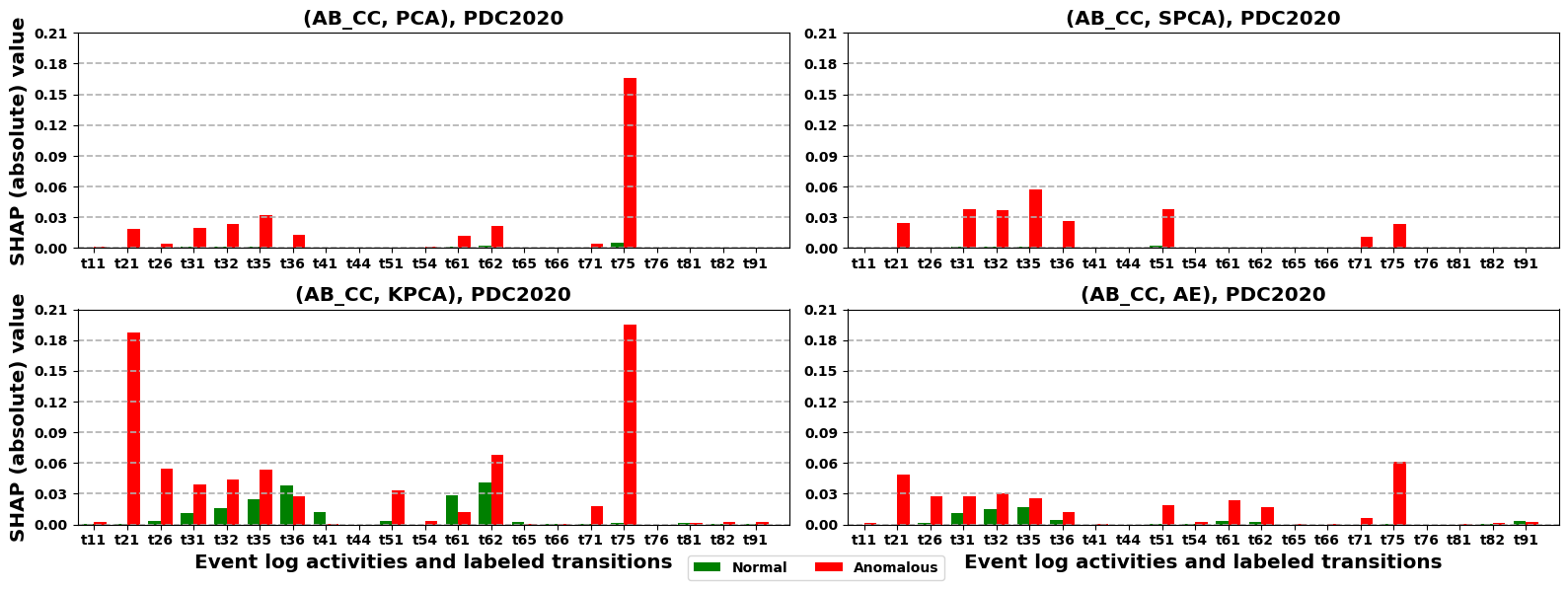}
\caption{SHAP (absoulte) values related to the additive feature-based explanation of the four \revision{conformance checking}-based framework techniques (AB\_CC,PCA), (AB\_CC,SPCA), (AB\_CC,KPCA), (AB\_CC,AE) using the PDC 2020 dataset.}
\label{SHAP_EXPLANATION}
\end{figure*}

\begin{figure*}[!t]
\centering
\includegraphics[width=\textwidth]{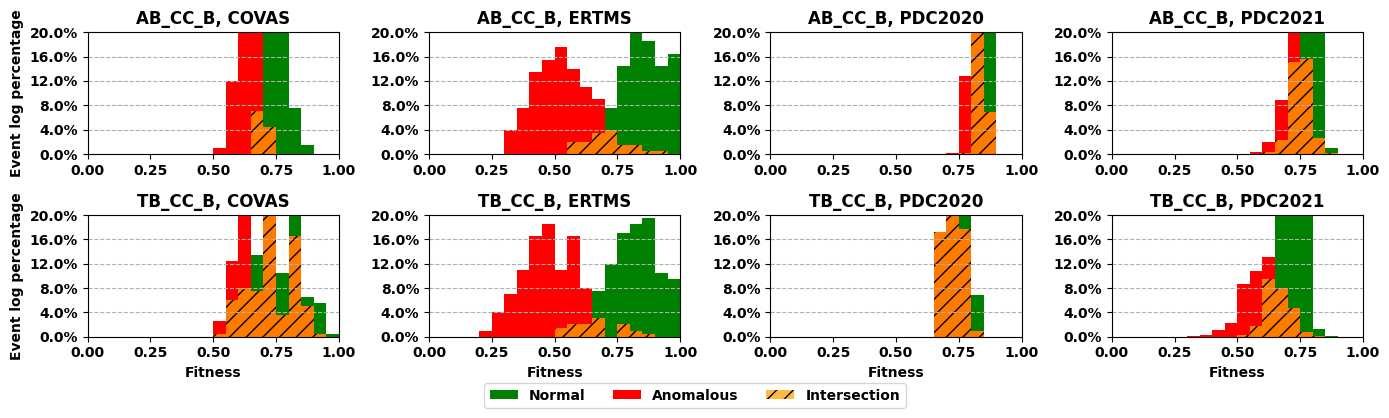}
\caption{The normal and anomalous histograms that show the frequency of occurrence of the fitness measures of normal and anomalous event logs per dataset for the AB\_CC\_B and TB\_CC\_B baseline approaches. The intersection highlights the overlap between anomalous and normal histograms.}
\label{FITNESS_VALUES_DISTRIBUTION}
\end{figure*}

\paragraph{Framework techniques explanation} Let us consider the anomaly explanation of techniques (AB\_CC, PCA), (AB\_CC, SPCA), (AB\_CC, KPCA), (AB\_CC, AE) for the PDC 2020 dataset, whose Petri net is depicted on the left side of Figure \ref{PETRI_NET_TRACES}. Figure \ref{SHAP_EXPLANATION} shows the SHAP (absolute) values for each technique. These values show the influence of the per-activity cost of each event log activity and labeled transition of the reference Petri net on the reconstruction error. For each activity, the mean SHAP value of anomalous (red) and normal (green) event logs is shown. The bar plot highlights that the high detection effectiveness of each framework technique is due to: 1) the small reconstruction error of per-activity costs in normal event logs; and 2) the high reconstruction error of per-activity costs in anomalous event logs. For example, the SHAP values of (AB\_CC, AE) highlight that the influence on reconstruction error of per-activity costs of normal event logs is strictly less than the reconstruction error of per-activity costs of anomalous event logs. In addition, each technique outlines that \texttt{t21} and \texttt{t75} influence the reconstruction error significantly, which suggests that the traces of anomalous event logs often lead to mismatching moves such as (\texttt{t21},$\nomove$), ($\nomove$,\texttt{t21}), which, in turn, could be due to \texttt{t21} being duplicated, skipped, or wrongly-ordered according to the Petri net's control-flow relations. In the case of \texttt{t75}, the activity is simply unknown, hence leading to the mismatching move (\texttt{t75}, $\nomove$) in the alignments.

\paragraph{Baselines explanation} Let us consider the distribution of fitness measures of normal and anomalous event logs for the AB\_CC\_B and TB\_CC\_B baselines. Figure \ref{FITNESS_VALUES_DISTRIBUTION} shows these distributions for each dataset, highlighting the intersection of the anomalous and normal distributions. These histograms provide insight into why the baseline approaches may fail at correctly classifying normal and anomalous event logs. In fact, the worst performance is obtained when there is excessive overlap of fitness values from normal and anomalous event logs, such as the PDC 2020/2021 datasets, where the four histograms of AB\_CC\_B and TB\_CC\_B show a significant overlap of normal and anomalous fitness values. Clearly, a higher-quality Petri net could provide much better results with the baselines. This is the case for the ERTMS dataset, where the anomalous and normal histograms of both the baselines are better separated, and, therefore, fitness thresholding is much more effective. Finally, the COVAS dataset leads to little overlap for AB\_CC\_B, whereas, for TB\_CC\_B, there is significant overlap. 
\begin{table*}[!t]
\centering
\caption{The mean $A$, $R$, $P$ and $F1$ and their standard deviation in subscripts for each framework and baseline technique and dataset. Measures in bold indicate the highest figure per dataset.}
\label{COMPARISON_RESULTS}
\resizebox{\textwidth}{!}{%
\begin{tabular}{llllllllllllllllllllll}
\toprule
\multicolumn{2}{l}{\textbf{}}          &  & \multicolumn{4}{l}{\textbf{PDC 2020}}                                                         &  & \multicolumn{4}{l}{\textbf{PDC 2021}}                                                         &  & \multicolumn{4}{l}{\textbf{ERTMS}}                                                            &  & \multicolumn{4}{l}{\textbf{COVAS}}                                                            \\ \cmidrule{4-7} \cmidrule{9-12} \cmidrule{14-17} \cmidrule{19-22} 
\multicolumn{2}{l}{\textbf{Technique}} &  & \textbf{A(\%)}        & \textbf{R(\%)}        & \textbf{P(\%)}        & \textbf{F1(\%)}       &  & \textbf{A(\%)}        & \textbf{R(\%)}        & \textbf{P(\%)}        & \textbf{F1(\%)}       &  & \textbf{A(\%)}        & \textbf{R(\%)}        & \textbf{P(\%)}        & \textbf{F1(\%)}       &  & \textbf{A(\%)}        & \textbf{R(\%)}        & \textbf{P(\%)}        & \textbf{F1(\%)}       \\ \midrule
\multicolumn{2}{l}{(AB\_CC, PCA)}      &  & 96.2$_{1.0}$          & 92.7$_{1.2}$          & 96.5$_{2.2}$          & 94.4$_{1.4}$          &  & 86.0$_{2.4}$          & 83.5$_{2.4}$          & 84.8$_{3.0}$          & 84.0$_{2.6}$          &  & 87.6$_{1.3}$          & 76.8$_{4.6}$          & 82.3$_{1.6}$          & 78.8$_{3.4}$          &  & 85.1$_{6.4}$          & 77.5$_{5.8}$          & 78.5$_{9.9}$          & 77.5$_{7.7}$          \\
\multicolumn{2}{l}{(AB\_CC, SPCA)}     &  & 96.8$_{1.0}$          & 93.8$_{1.4}$          & 97.1$_{2.0}$          & 95.3$_{1.4}$          &  & 86.3$_{2.9}$          & 83.1$_{2.6}$          & 85.8$_{3.8}$          & 84.1$_{3.1}$          &  & 88.2$_{2.3}$          & 78.8$_{5.5}$          & 82.5$_{3.4}$          & 80.3$_{4.7}$          &  & 91.0$_{4.8}$          & 87.6$_{7.2}$          & 85.8$_{7.5}$          & 86.5$_{7.0}$          \\
\multicolumn{2}{l}{(AB\_CC, KPCA)}     &  & 96.4$_{0.7}$          & 92.1$_{1.5}$          & 97.8$_{0.4}$          & 94.6$_{1.1}$          &  & 86.6$_{1.4}$          & 83.5$_{2.3}$          & 85.9$_{1.6}$          & 84.4$_{1.9}$          &  & 89.3$_{1.7}$          & 87.8$_{2.3}$          & 83.1$_{3.4}$          & 84.7$_{1.7}$          &  & \textbf{92.8$_{1.0}$} & \textbf{87.9$_{3.2}$} & \textbf{89.5$_{1.1}$} & \textbf{88.5$_{2.0}$} \\
\multicolumn{2}{l}{(AB\_CC, AE)}       &  & \textbf{98.1$_{0.3}$} & \textbf{96.6$_{1.3}$} & \textbf{98.1$_{0.5}$} & \textbf{97.3$_{0.5}$} &  & 88.9$_{1.1}$          & 86.2$_{1.3}$          & 88.4$_{1.4}$          & 87.2$_{1.3}$          &  & 91.2$_{1.9}$          & 83.3$_{5.8}$          & 88.3$_{1.6}$          & 85.2$_{4.1}$          &  & 91.4$_{3.1}$          & 85.6$_{6.5}$          & 87.0$_{5.0}$          & 86.2$_{5.7}$          \\ \midrule
\multicolumn{2}{l}{(TB\_CC, PCA)}      &  & 96.1$_{0.3}$          & 91.7$_{3.1}$          & 97.5$_{0.2}$          & 94.1$_{0.5}$          &  & 90.0$_{0.6}$          & 86.6$_{0.6}$          & 90.7$_{1.0}$          & 88.2$_{0.7}$          &  & 89.6$_{2.1}$          & 75.6$_{5.3}$          & 90.6$_{1.7}$          & 80.0$_{5.1}$          &  & 52.1$_{3.0}$          & 57.9$_{3.0}$          & 55.2$_{2.1}$          & 48.6$_{0.9}$          \\
\multicolumn{2}{l}{(TB\_CC, SPCA)}     &  & 96.5$_{0.7}$          & 92.9$_{1.4}$          & 97.2$_{1.5}$          & 94.8$_{1.1}$          &  & \textbf{90.6$_{0.9}$} & \textbf{87.1$_{1.3}$} & \textbf{91.7$_{0.8}$} & \textbf{88.9$_{1.2}$} &  & 92.6$_{1.0}$          & 85.5$_{3.6}$          & 90.9$_{1.7}$          & 87.6$_{2.3}$          &  & 55.6$_{1.7}$          & 61.2$_{1.3}$          & 57.2$_{0.8}$          & 52.0$_{1.2}$          \\
\multicolumn{2}{l}{(TB\_CC, KPCA)}     &  & 96.5$_{1.4}$          & 92.6$_{3.1}$          & 97.5$_{1.0}$          & 94.7$_{2.3}$          &  & 67.0$_{4.8}$          & 70.2$_{3.1}$          & 68.2$_{2.4}$          & 66.3$_{4.4}$          &  & 92.4$_{1.4}$          & 88.6$_{1.6}$          & 88.1$_{2.6}$          & 88.2$_{1.7}$          &  & 40.8$_{5.5}$          & 53.1$_{3.2}$          & 52.4$_{2.6}$          & 39.6$_{3.9}$          \\
\multicolumn{2}{l}{(TB\_CC, AE)}       &  & 96.6$_{0.4}$          & 92.6$_{1.0}$          & 97.9$_{0.2}$          & 95.0$_{0.7}$          &  & 90.0$_{0.6}$          & 87.3$_{1.8}$          & 91.3$_{0.8}$          & 88.8$_{1.5}$          &  & \textbf{94.0$_{0.5}$} & \textbf{89.8$_{2.7}$} & \textbf{91.5$_{1.3}$} & \textbf{90.5$_{1.0}$} &  & 73.8$_{3.1}$          & 69.7$_{2.8}$          & 64.8$_{2.3}$          & 65.7$_{2.7}$          \\ \midrule
\multicolumn{2}{l}{(DF, PCA)}          &  & 58.6$_{5.6}$          & 60.4$_{3.4}$          & 57.4$_{2.4}$          & 54.5$_{4.2}$          &  & 79.4$_{1.6}$          & 77.1$_{1.4}$          & 77.0$_{1.8}$          & 76.9$_{1.5}$          &  & 89.2$_{1.1}$          & 76.9$_{3.8}$          & 87.1$_{1.6}$          & 80.4$_{3.4}$          &  & 89.5$_{2.0}$          & 78.6$_{5.4}$          & 86.6$_{2.4}$          & 81.4$_{4.5}$          \\
\multicolumn{2}{l}{(DF, SPCA)}         &  & 62.2$_{2.8}$          & 62.4$_{0.4}$          & 58.9$_{0.4}$          & 57.2$_{1.5}$          &  & 80.4$_{1.7}$          & 78.6$_{1.2}$          & 78.3$_{1.9}$          & 78.2$_{1.3}$          &  & 88.4$_{2.0}$          & 74.8$_{5.7}$          & 86.7$_{1.8}$          & 78.4$_{5.0}$          &  & 89.2$_{1.1}$          & 78.4$_{4.3}$          & 85.8$_{0.3}$          & 81.1$_{3.1}$          \\
\multicolumn{2}{l}{(DF, KPCA)}         &  & 70.3$_{2.2}$          & 67.9$_{1.7}$          & 63.7$_{1.5}$          & 64.1$_{1.9}$          &  & 78.4$_{0.9}$          & 76.6$_{0.8}$          & 75.9$_{1.0}$          & 76.1$_{0.6}$          &  & 89.2$_{1.1}$          & 77.2$_{3.2}$          & 86.8$_{1.3}$          & 80.6$_{2.8}$          &  & 91.9$_{0.0}$          & 83.5$_{1.2}$          & 90.1$_{1.8}$          & 86.2$_{1.3}$          \\
\multicolumn{2}{l}{(DF, AE)}           &  & 71.6$_{2.2}$          & 69.3$_{0.0}$          & 64.8$_{0.0}$          & 65.4$_{0.0}$          &  & 70.1$_{0.0}$          & 72.3$_{0.0}$          & 69.8$_{0.0}$          & 69.2$_{0.0}$          &  & 89.7$_{1.3}$          & 80.2$_{2.4}$          & 86.0$_{2.6}$          & 82.6$_{2.4}$          &  & 90.8$_{1.4}$          & 83.9$_{3.2}$          & 86.4$_{2.1}$          & 85.0$_{2.7}$          \\ \midrule
\multicolumn{2}{l}{(NG, PCA)}          &  & 84.0$_{1.0}$          & 75.5$_{2.1}$          & 77.3$_{1.6}$          & 76.3$_{1.8}$          &  & 75.0$_{2.5}$          & 73.5$_{2.2}$          & 72.3$_{2.4}$          & 72.7$_{2.4}$          &  & 87.6$_{1.1}$          & 77.8$_{3.7}$          & 81.6$_{1.5}$          & 79.3$_{2.6}$          &  & 91.3$_{0.8}$          & 83.5$_{1.4}$          & 88.3$_{1.7}$          & 85.6$_{4.0}$          \\
\multicolumn{2}{l}{(NG, SPCA)}         &  & 83.8$_{0.0}$          & 76.0$_{1.7}$          & 77.1$_{1.0}$          & 76.5$_{1.1}$          &  & 75.3$_{0.9}$          & 73.5$_{1.8}$          & 72.6$_{1.0}$          & 72.8$_{1.3}$          &  & 87.6$_{2.0}$          & 76.8$_{6.8}$          & 82.1$_{2.2}$          & 78.5$_{5.4}$          &  & 87.6$_{0.8}$          & 75.3$_{1.5}$          & 83.0$_{1.7}$          & 78.1$_{1.5}$          \\
\multicolumn{2}{l}{(NG, KPCA)}         &  & 85.6$_{0.7}$          & 75.7$_{3.1}$          & 80.9$_{0.9}$          & 77.6$_{2.0}$          &  & 72.9$_{2.5}$          & 73.1$_{1.6}$          & 71.1$_{1.4}$          & 71.2$_{1.6}$          &  & 87.6$_{1.4}$          & 75.3$_{1.1}$          & 83.2$_{3.9}$          & 78.2$_{2.0}$          &  & 92.2$_{2.1}$          & 86.0$_{5.6}$          & 89.1$_{2.5}$          & 87.2$_{4.0}$          \\
\multicolumn{2}{l}{(NG, AE)}           &  & 87.2$_{0.0}$          & 78.2$_{0.0}$          & 83.3$_{0.0}$          & 80.3$_{0.0}$          &  & 77.0$_{0.0}$          & 75.3$_{0.0}$          & 74.3$_{0.0}$          & 74.7$_{0.0}$          &  & 86.0$_{0.8}$          & 73.1$_{2.3}$          & 79.5$_{1.4}$          & 75.5$_{2.0}$          &  & 92.0$_{0.0}$          & 83.7$_{0.0}$          & 90.1$_{0.0}$          & 86.4$_{0.0}$          \\ \bottomrule
\end{tabular}%
}
\end{table*}
\begin{figure*}[t]
\centering
\includegraphics[width=\textwidth]{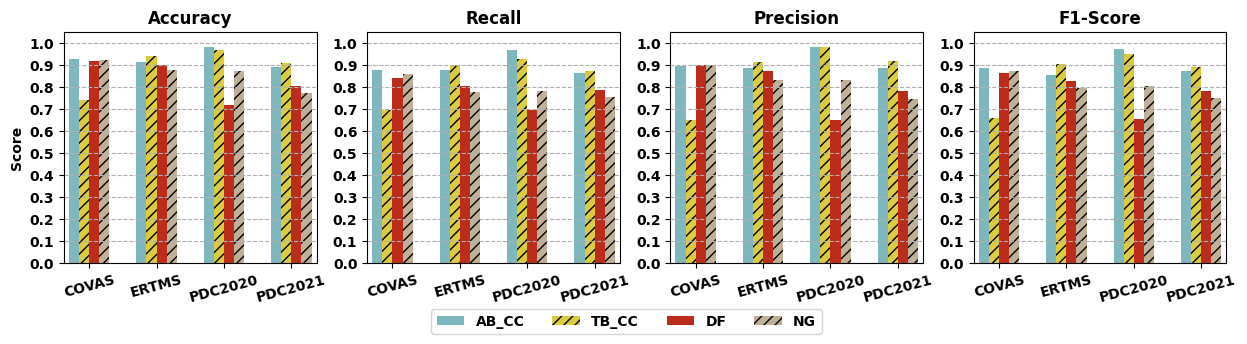}
\caption{The mean $A$, $R$, $P$ and $F1$ per dataset of the best-performing dimensionality reduction technique for each \revision{process mining}-based feature extraction approach.}
\label{BAR_PLOT_COMPARISON}
\end{figure*}

\subsubsection{Framework techniques comparisons}
In this part, we compare the detection effectiveness of framework techniques using all four \revision{process mining}-based feature extraction approaches. Each technique is applied to each dataset 5 times to account for statistical fluctuations due to the random split of training, validation and test sets.

\paragraph{Results}
Table \ref{COMPARISON_RESULTS} reports the mean $A$, $R$, $P$ and $F1$ measures and their standard deviation in subscripts for each technique and dataset. For the PDC 2020 and COVAS datasets, (AB\_CC, AE) and (AB\_CC, KPCA) show the best performance, achieving 97.3\% and 88.5\% $F1$. For the PDC 2021 and ERTMS datasets, (TB\_CC, SPCA) and (TB\_CC, AE) show the best performance, achieving 88.9\% and 90.5\% $F1$. Arguably, TB\_CC \revision{process mining}-based feature extraction is better than AB\_CC for the PDC 2021 and ERTMS dataset because of the explanation in Figure \ref{FITNESS_VALUES_DISTRIBUTION}. From this figure, the fitness measures of the TB\_CC\_B baseline with respect to the ERTMS dataset are better separated than those fitness measures of the AB\_CC\_B baseline. Hence, the TB\_CC approach could have led to better features than AB\_CC. Figure \ref{BAR_PLOT_COMPARISON} depicts the bar plots that capture the impact of the different \revision{process mining}-based feature extraction approaches on the performance measures for each dataset. In particular, given a PF value, the dimensionality reduction technique providing the best performance is considered to compare the best-performing PF-DR value pairs. For example, given the PDC 2020 dataset, the best-performing PF-DR value pairs are (AB\_CC, AE), (TB\_CC, AE), (DF, AE) and (NG, AE), with $F1$ equal to 97.3\%, 95.0\%, 65.4\% and 80.3\%. 

\paragraph{Analysis of variance}
We conduct an analysis of variance to evaluate whether the dataset, the \revision{process mining}-based feature extraction approach, and the two combined result in statistically significant differences. To this aim, we design a two-factor full factorial experiment with factors PF and DS, of which the latter is the dataset factor. In like manner as Figure \ref{BAR_PLOT_COMPARISON}, for each PF value, we take the best-performing dimensionality reduction technique for the given dataset. Table \ref{ANOVA_COMPARISON} reports the mean values of the performance measures for each DS-PF value pair, specifying whether the DS factor, PF factor, and DS and PF factors combined are significant for the target metric using the non-parametric Friedman test (a p-value less than 0.05 implies that the factor is significant with 95\% confidence). In addition, the table shows how important the aforementioned factors are in terms of explained variance. Notably, the proposed AB\_CC \revision{process mining}-based feature extraction approach shows satisfying performance in all cases, e.g., $F1$ drops only as low as 85.2\% for the ERTMS dataset and peaks at 97.3\% for the PDC 2020 dataset. However, it is worth noting that AB\_CC is outperformed by by TB\_CC for the ERTMS and PDC 2021 datasets. Therefore, it is safe to state that there is no one-size-fits-all \revision{process mining}-based feature extraction approach valid for all datasets; besides, this is also confirmed statistically by the analysis of variance, which outlines that the interaction between the DS and PF factors is significant and important for all the measures considered.
\begin{table*}[!t]
\centering
\caption{The mean $A$, $R$, $P$ and $F1$ measures and their standard deviation in subscripts per dataset (DS) for each \revision{process mining}-based feature extraction (PF) approach. Measures in bold indicate the highest figure per dataset. At the bottom is the significance of the DS and PF factors per metric $X$, where $p_X$ indicates the p-value of the Friedman test ($p_X<0.05$ means the factor is significant with 95\% confidence) and $I_X$ the percentage of variance in $X$ explained by the factor.}
\label{ANOVA_COMPARISON}
\resizebox{\textwidth}{!}{%
\begin{tabular}{llllllllllllllllllll}
\toprule
\multirow{2}{*}{\textbf{\diagbox[]{DS}{PF}}} & \multicolumn{4}{l}{\textbf{AB\_CC}}                                                           &  & \multicolumn{4}{l}{\textbf{TB\_CC}}                                                           &  & \multicolumn{4}{l}{\textbf{DF}}                                    &  & \multicolumn{4}{l}{\textbf{NG}}                                    \\ \cmidrule{2-5} \cmidrule{7-10} \cmidrule{12-15} \cmidrule{17-20} 
                                & \textbf{A(\%)}        & \textbf{R(\%)}        & \textbf{P(\%)}        & \textbf{F1(\%)}       &  & \textbf{A(\%)}        & \textbf{R(\%)}        & \textbf{P(\%)}        & \textbf{F1(\%)}       &  & \textbf{A(\%)} & \textbf{R(\%)} & \textbf{P(\%)} & \textbf{F1(\%)} &  & \textbf{A(\%)} & \textbf{R(\%)} & \textbf{P(\%)} & \textbf{F1(\%)} \\ \midrule
\textbf{PDC 2020}               & \textbf{98.1$_{0.3}$} & \textbf{96.6$_{1.3}$} & \textbf{98.1$_{0.5}$} & \textbf{97.3$_{0.5}$} &  & 96.6$_{0.4}$          & 92.9$_{1.4}$          & 97.9$_{0.2}$          & 95.0$_{0.7}$          &  & 71.6$_{2.2}$   & 69.3$_{0.0}$   & 64.8$_{0.0}$   & 65.4$_{0.0}$    &  & 87.2$_{0.0}$   & 78.2$_{0.0}$   & 83.3$_{0.0}$   & 80.3$_{0.0}$    \\
\textbf{PDC 2021}               & 88.9$_{1.1}$          & 86.2$_{1.3}$          & 88.4$_{1.4}$          & 87.2$_{1.3}$          &  & \textbf{90.6$_{0.9}$} & \textbf{87.1$_{1.3}$} & \textbf{91.7$_{0.8}$} & \textbf{88.9$_{1.2}$} &  & 80.4$_{1.7}$   & 78.6$_{1.2}$   & 78.3$_{1.9}$   & 78.2$_{1.3}$    &  & 77.0$_{0.0}$   & 75.3$_{0.0}$   & 74.3$_{0.0}$   & 74.7$_{0.0}$    \\
\textbf{ERTMS}                  & 91.2$_{1.9}$          & 83.3$_{5.8}$          & 88.3$_{1.6}$          & 85.2$_{4.1}$          &  & \textbf{94.0$_{0.5}$} & \textbf{89.8$_{2.7}$} & \textbf{91.5$_{1.3}$} & \textbf{90.5$_{1.0}$} &  & 89.7$_{1.3}$   & 80.2$_{2.4}$   & 86.0$_{2.6}$   & 82.6$_{2.4}$    &  & 87.6$_{1.1}$   & 77.8$_{3.7}$   & 81.6$_{1.5}$   & 79.3$_{2.6}$    \\
\textbf{COVAS}                  & \textbf{92.8$_{1.0}$} & \textbf{87.9$_{3.2}$} & \textbf{89.5$_{1.1}$} & \textbf{88.5$_{2.0}$} &  & 73.8$_{3.1}$          & 69.7$_{2.8}$          & 64.8$_{2.3}$          & 65.7$_{2.7}$          &  & 91.9$_{0.0}$   & 83.5$_{1.2}$   & 90.1$_{1.8}$   & 86.2$_{1.3}$    &  & 92.2$_{2.1}$   & 86.0$_{5.6}$   & 89.1$_{2.5}$   & 87.2$_{4.0}$    \\ \midrule
\multicolumn{20}{l}{\textbf{DS significance and importance: $p_A=0.00, I_A=8.7\%; p_R=0.13, I_R=2.0\%; p_P=0.05, I_P=3.2\%; p_{F1}=0.02, I_{F1}=2.6\%$}}                                                                                                                                                                                                                           \\
\multicolumn{20}{l}{\textbf{PF significance and importance: $p_A=0.00, I_A=20.0\%; p_R=0.00, I_R=34.3\%; p_P=0.00, I_P=17.8\%; p_{F1}=0.00, I_{F1}=25.7\%$}}                                                                                                                                                                                                                       \\
\multicolumn{20}{l}{\textbf{DS and PF significance and importance: $p_A=0.00, I_A=68.7\%; p_R=0.00, I_R=55.2\%; p_P=0.00, I_P=76.8\%; p_{F1}=0.00, I_{F1}=66.0\%$}}                                                                                                                                                                                                                \\ \bottomrule
\end{tabular}%
}
\end{table*}

\section{Conclusions}
\label{sec:conclusions}
In this paper, we took on the detection of control-flow anomalies in the business processes of organizations. To address the shortcomings of existing \revision{conformance checking}-based techniques and support the development of competitive techniques for control-flow anomaly detection while maintaining the explainable nature of \revision{conformance checking}, we proposed a novel \revision{process mining}-based feature extraction approach with alignment-based \revision{conformance checking}. This \revision{conformance checking} variant aligns the deviating control flow with a reference process model; the resulting alignment can be inspected to extract additional statistics such as the number of times a given activity caused mismatches. Secondly, we integrated this approach into a flexible and explainable framework for developing techniques for control-flow anomaly detection. The framework combines \revision{process mining}-based feature extraction and dimensionality reduction, which can handle high-dimensional feature sets, achieve detection effectiveness, and support explainability. In addition to our proposed \revision{process mining}-based feature extraction approach, the framework allows employing other approaches, enabling flexible development of multiple \revision{conformance checking}-based and \revision{conformance checking}-independent techniques for control-flow anomaly detection. Using a set of synthetic and real-world datasets, we experimented with several \revision{conformance checking}-based and \revision{conformance checking}-independent framework techniques, and compared the results with baseline \revision{conformance checking}-based techniques. The results showed that the \revision{conformance checking}-based framework techniques using our approach outperform the baseline \revision{conformance checking}-based techniques while maintaining the explainable nature of \revision{conformance checking}. In particular, the best-performing framework techniques achieved 97.3\%, 88.9\%, 90.5\% and 88.5\% $F1$ for the PDC 2020, PDC 2021, ERTMS and COVAS datasets, whereas the baseline \revision{conformance checking}-based techniques could achieve up to 36.1\%, 56.6\%, 78.8\%, and 79.1\%. We also provide an explanation of why existing \revision{conformance checking}-based techniques may be ineffective. Finally, the results indicated that detection effectiveness is not solely dependent on the specific framework techniques used, as no one-size-fits-all \revision{process mining}-based feature extraction approach is suitable for all the datasets. In fact, analysis of variance outlined that both the dataset and \revision{process mining}-based feature extraction approach show statistically significant differences and high importance in $F1$ values.

Future work will extend the capabilities of the framework by considering additional event log attributes including information related to other perspectives as well as the control-flow one \cite{aalst2022pmhandbook}. Such information can be used in \revision{process mining}-based feature extraction to enrich the tabular data, address other types of anomalies, and result in improved detection effectiveness. The extension will also include object-centric process mining, which allows for focusing on the activities of the different entities involved in a process instead of flattening the event data and treating all objects uniformly \cite{aalst2023ocpm}.

\section*{Acknowledgement}
This work has been partially supported by the Spoke 9 “Digital Society \& Smart Cities” of ICSC - Centro Nazionale di Ricerca in High Performance-Computing, Big Data and Quantum Computing, funded by the European Union - NextGenerationEU (PNRR-HPC, CUP: E63C22000980007).

\bibliographystyle{elsarticle-harv} 
\bibliography{bibliography.bib}

\end{document}